\documentclass[10pt, a2paper]{article}

\usepackage[utf8]{inputenc}
\usepackage[T1]{fontenc}
\usepackage[numbers]{natbib}
\usepackage{tcolorbox}
\usepackage{listings}
\usepackage{tcolorbox}
\tcbuselibrary{listings,breakable,skins} 
\usepackage{geometry}
\usepackage{graphicx}
\usepackage{subcaption}
\usepackage{natbib}
\usepackage{float}
\usepackage{booktabs}   
\usepackage{longtable}  
\usepackage{enumitem}   
\usepackage{array}      
\usepackage{ragged2e}   
\usepackage{amsmath}
\usepackage{amsfonts}
\usepackage{adjustbox}
\usepackage{tabularx}
\usepackage[table]{xcolor}
\usepackage{csquotes}
\usepackage{pifont}     
\usepackage{CJKutf8} 
\usepackage{parskip} 
\usepackage{titlesec} 
\usepackage{hyperref}
\usepackage{tikz}
\usepackage{xltabular}
\usepackage{eso-pic}

\hypersetup{
  colorlinks=true,
  urlcolor=blue
}
\titlespacing*{\paragraph}{0pt}{1.5ex plus 0.5ex minus .2ex}{1em}
\usepackage{enumitem}
\setlist[itemize]{
    itemsep=4pt,      
    parsep=0pt,       
    topsep=6pt,       
    partopsep=0pt,    
    leftmargin=* 
}

\definecolor{headerbg}{RGB}{240, 242, 245}
\definecolor{goodgreen}{RGB}{40, 140, 60}
\definecolor{badred}{RGB}{200, 50, 50}
\definecolor{principlebg}{RGB}{248, 249, 250} 

\newcommand{\cmark}{\textcolor{goodgreen}{\ding{51}}}
\newcommand{\xmark}{\textcolor{badred}{\ding{55}}}

\newcommand{\PlaceFirstPageLogo}{%
  \AddToShipoutPictureFG*{%
    \put(\LenToUnit{\dimexpr1in+\oddsidemargin\relax},
         \LenToUnit{\dimexpr\paperheight-1.2cm\relax}){%
      \makebox(0,0)[lt]{%
        \includegraphics[width=2.8cm]{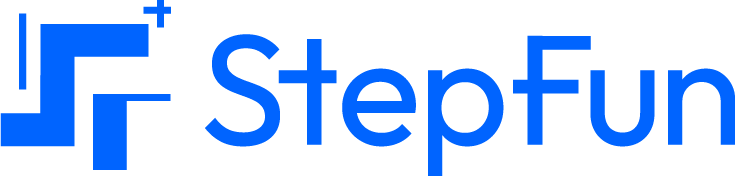}%
      }%
    }%
    \put(\LenToUnit{\dimexpr1in+\oddsidemargin+\textwidth\relax},
         \LenToUnit{\dimexpr\paperheight-1.6cm\relax}){%
      \makebox(0,0)[rt]{\small 2025-12-24}%
    }%
    \put(\LenToUnit{\dimexpr1in+\oddsidemargin\relax},
         \LenToUnit{\dimexpr\paperheight-2.2cm\relax}){%
      \makebox(0,0)[lt]{\rule{\textwidth}{0.4pt}}%
    }%
  }%
}

\title{\textbf{Step-DeepResearch Technical Report}}
\vspace{-2em}
\author{\textbf{Agent-Team}, \textbf{StepFun}}
\date{\vspace{-1.5em}}
\begin{document}

\PlaceFirstPageLogo   

\maketitle
\begin{center}
\url{https://github.com/stepfun-ai/StepDeepResearch}
\end{center}
\begin{abstract}
\noindent As Large Language Models (LLMs) shift toward autonomous agents, Deep Research has emerged as a pivotal metric for assessing the core competitiveness of agents. However, existing works primarily focus on academic multi-hop search tasks with ground truth, such as BrowseComp, which often struggle to satisfy user demands for open-ended research tasks in real-world scenarios. Open-ended research not only demands robust retrieval capabilities but also challenges the agent's comprehensive skills in latent intent recognition, long-horizon decision-making, multi-turn tool use, logical structuring, and cross-source verification. To address this, we introduce \textbf{Step-DeepResearch}, a cost-effective, end-to-end Deep Research agent model. We propose a novel \textit{Data Synthesis Strategy Based on Atomic Capabilities}, aimed at reinforcing underlying capabilities in planning, information seeking, reflection, and report writing. In terms of the training paradigm, we construct a progressive path from agentic mid-training to supervised fine-tuning and reinforcement learning. Combined with a \textit{Checklist-style Judger} reward design, this approach significantly improves robustness across diverse scenarios. Furthermore, to address the lack of evaluations reflecting real-world demands in the Chinese domain, we establish \textbf{ADR-Bench}, a Chinese benchmark for realistic Deep Research scenarios. Experimental results demonstrate that Step-DeepResearch, with only 32B parameters, achieves a high score of \textbf{61.42} on the Scale AI \textsc{ResearchRubrics}. In expert human evaluations on ADR-Bench, its Elo score significantly outperforms comparable models and rivals state-of-the-art proprietary services such as OpenAI DeepResearch and Gemini DeepResearch. These findings prove that through a refined training scheme, medium-sized models can achieve expert-level Deep Research capabilities. With extremely low deployment and inference costs, Step-DeepResearch stands as the most cost-effective Deep Research agent model currently available in the industry.
\end{abstract}
\vspace{+1.5em}

    

\begin{figure}[htbp]
    \centering
    \begin{subfigure}[b]{0.64\textwidth}
        \centering
        \includegraphics[width=1\linewidth]{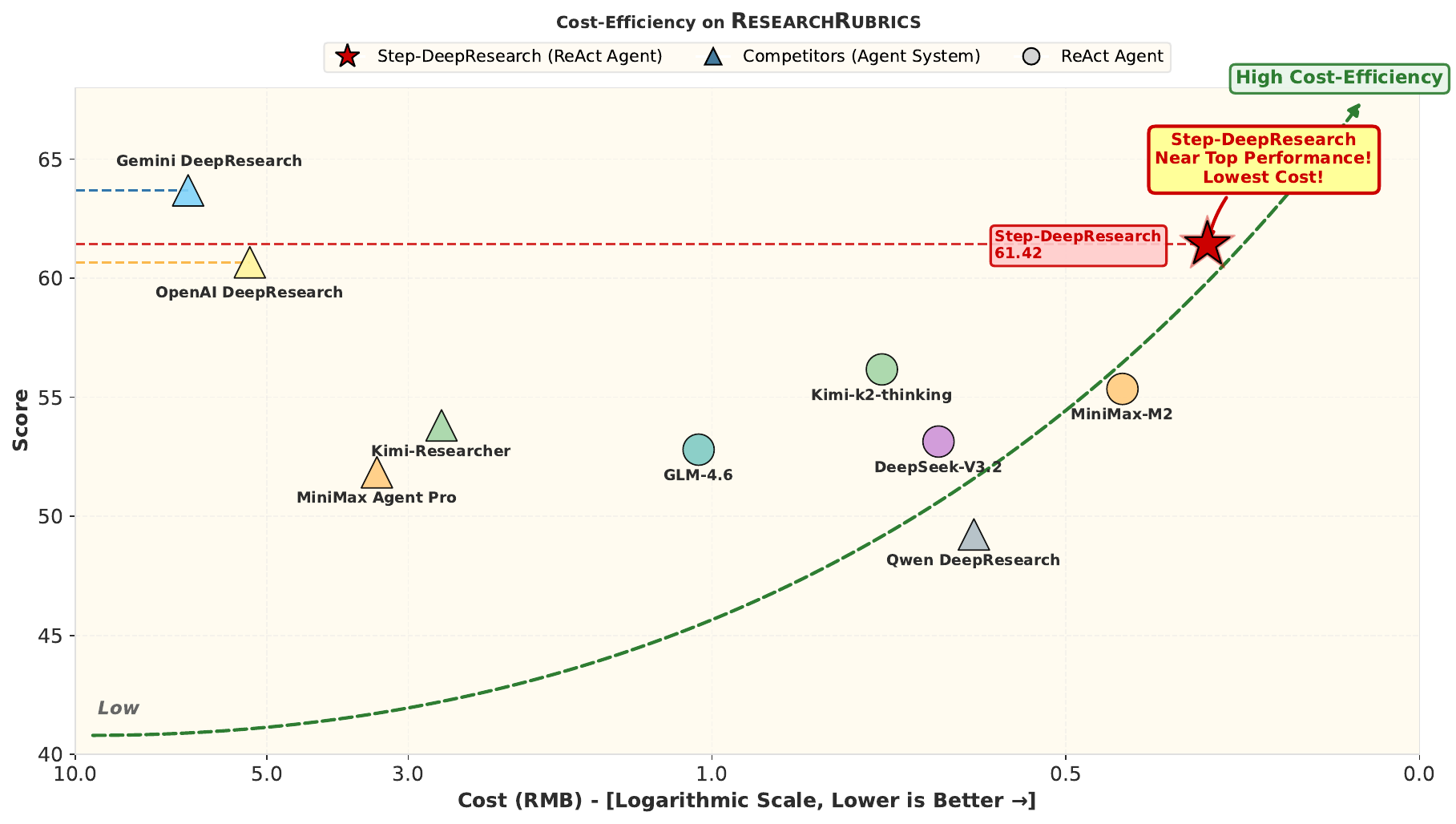}
        \caption{}
        \label{fig:cost_perf}
    \end{subfigure}
    \hspace{0.02\textwidth}
    \begin{subfigure}[b]{0.32\textwidth}
        \centering
        \includegraphics[width=1.08\linewidth]{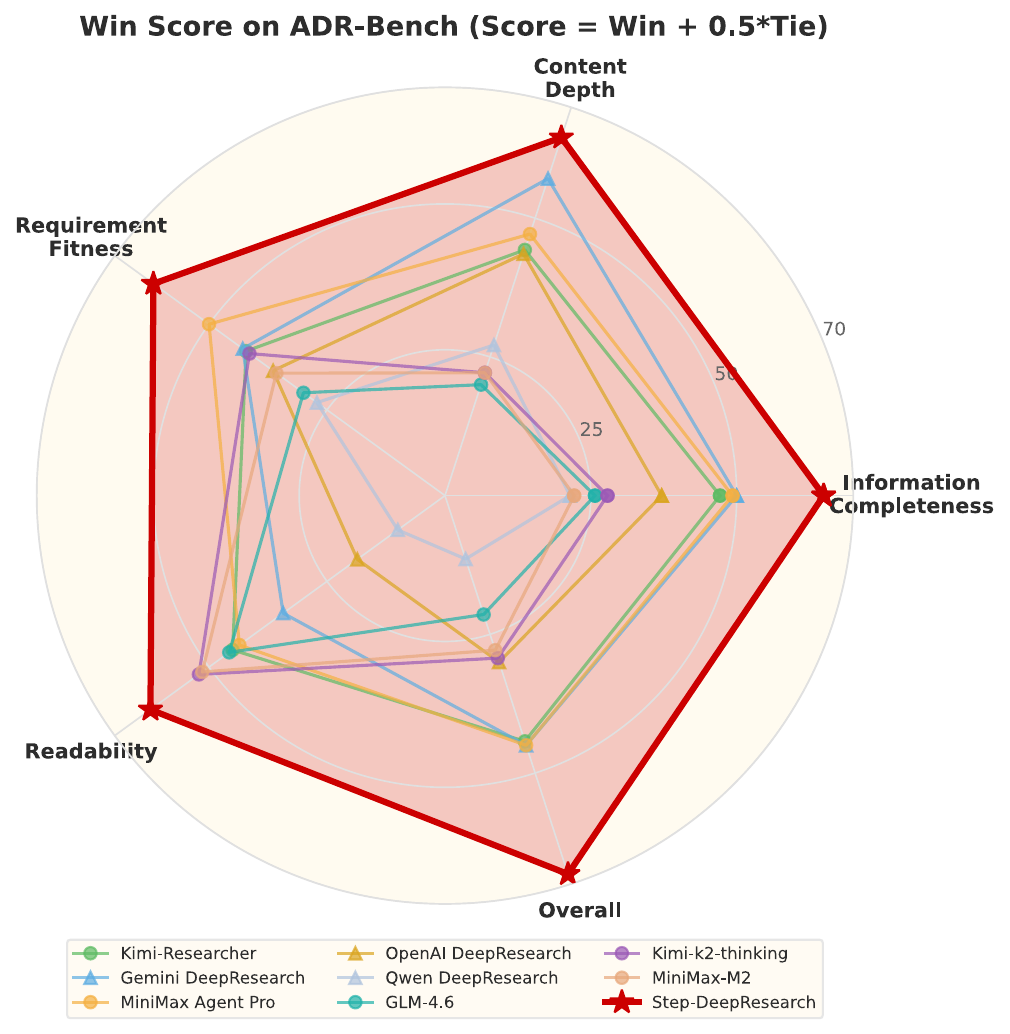}
        \caption{}
        \label{fig:elo_scores}
    \end{subfigure}

    \caption{\textbf{Comprehensive Evaluation of Step-DeepResearch.} 
    (a) \textbf{Cost-Efficiency on \textsc{ResearchRubrics}:} Step-DeepResearch achieves near-top performance (61.42) while significantly reducing inference costs (RMB), positioned at the high-efficiency frontier. 
    (b) \textbf{Expert Evaluation on ADR-Bench:} Step-DeepResearch consistently leads in Elo ratings across all dimensions, rivaling top-tier closed-source models.}
    \label{fig:total_eval}
\end{figure}

\section{Introduction}
Large Language Models (LLMs) are rapidly pushing AI systems beyond conversational interaction toward autonomous agents. In this broader shift, \textit{Deep Research}, defined as the ability of AI systems to address open-ended, long-horizon, and highly complex information-seeking tasks, has become a useful ``North Star'' capability for general-purpose agents. Recent systems such as OpenAI DeepResearch~\cite{openai2025deepresearch}, Gemini DeepResearch~\cite{google2024gemini}, and Kimi-Researcher~\cite{kimi2025researcher} illustrate the potential of agentic information acquisition, while also revealing clear limitations.

\paragraph{Search is not research.}
Despite rapid industrial progress, there remains a substantial gap between benchmark performance and real-world usefulness. A core reason is a mismatch in task formulation: \textit{search is not equivalent to research}. Search typically targets well-specified queries with closed-form answers and optimizes for retrieval accuracy. Research, in contrast, is an iterative process that requires intent decomposition, planning, effective tool use, cross-source verification, and synthesis into a structured report.

\paragraph{Limits of multi-hop QA as an evaluation driver.}
In practice, multi-hop question answering is a common lens for evaluating agent systems, and it often becomes the de facto optimization target. This pressure biases agents toward retrieval-heavy behavior. They look more like efficient \emph{web crawlers} that collect scattered facts at scale, rather than \emph{researchers} who integrate evidence into a coherent and defensible argument. As a result, information fragmentation, broken reasoning chains, and hallucinations under noisy evidence remain major obstacles to deploying Deep Research systems in practice.

\paragraph{Our perspective.}
We reframe Deep Research as long-horizon decision-making over a set of atomic capabilities, including adaptive planning, information gathering and cross-source verification, reflection and error correction, and creative writing. From this perspective, improving usability is less about assembling external components, and more about training models to internalize an expert-like cognitive loop, so they can self-check and revise as the task unfolds.

\paragraph{Step-DeepResearch.}
Guided by these principles, we introduce \textbf{Step-DeepResearch}, an end-to-end framework that aims to enable robust and practical autonomous research primarily through native model capabilities. Our contributions are:

\begin{itemize}
    \item \textbf{Atomic-capability data synthesis.}
    We decompose Deep Research into trainable atomic capabilities and synthesize targeted data for agentic mid-training, covering domain knowledge, high-level planning, behavioral reflection, information summarization, and cross-source verification. To mitigate the scarcity of high-value reasoning data, we further develop a post-training data synthesis pipeline grounded in knowledge graphs and expert trajectories. This design aims to improve the information density and logical structure of training data, and to reduce missing-capability issues in conventional synthetic datasets.
    \item \textbf{Progressive training pipeline.}
    We establish a practical optimization path from agentic mid-training to supervised fine-tuning, and then to reinforcement learning. Using atomic-capability data, we reshape the objective from ``predicting the next token'' to ``deciding the next atomic action,'' which empirically improves robustness in complex environments and strengthens generalization across tasks.
    \item \textbf{Application-driven evaluation suite.}
    While existing benchmarks (e.g., BrowseComp~\cite{wei2025browsecomp} and GAIA~\cite{mialon2023gaia}) are informative, they often fall short of capturing real user needs. Moreover, high-quality Chinese benchmarks for Deep Research remain limited. We therefore build \textbf{ADR-Bench} (Application-driven Deep Research Benchmark), spanning commercial research, policy analysis, and software engineering. ADR-Bench combines an Elo-style rating protocol with multi-dimensional quality criteria to better connect automated metrics with human-perceived usefulness.
\end{itemize}

With 32B parameters, Step-DeepResearch shows strong performance relative to its scale. On ADR-Bench, it achieves high usability in end-to-end research. Unlike many Deep Research systems that rely on complex multi-agent coordination or heavyweight workflows, Step-DeepResearch benefits from internalized atomic capabilities, using only a streamlined ReAct-style single-agent design. In expert-based Elo ratings, it not only outperforms larger models like MiniMax-M2~\cite{minimaxm2}, GLM-4.6~\cite{zeng2025glm45}, and DeepSeek-V3.2~\cite{deepseek2025v32}, but also surpasses cutting-edge Deep Research systems such as Kimi-Researcher~\cite{kimi2025researcher} and MiniMax Agent Pro~\cite{minimaxm2}. On the \textsc{ResearchRubrics} Benchmark~\cite{sharma2025researchrubrics}, Step-DeepResearch achieves 61.42 rubric compliance under ternary grading, performing at a level comparable to OpenAI DeepResearch~\cite{openai2025deepresearch} and Gemini DeepResearch~\cite{google2024gemini}, and significantly outperforming a range of open-source and proprietary models. This work demonstrates that models with medium-scale parameters can also achieve expert-level Deep Research capabilities. As shown in Figure~\ref{fig:total_eval}, with its low deployment and inference costs, Step-DeepResearch has become the most cost-effective Deep Research system.

\section{Related Work}
\subsection{ Workflow-based Implementation of Deep Research}
Current mainstream Deep Research systems employ general foundation models, achieving research capabilities through external workflow orchestration. 
OpenAI DeepResearch~\cite{openai2025deepresearch} combines the o3-mini reasoning model with multi-step web exploration, adopting asynchronous task management. 
Gemini DeepResearch~\cite{google2024gemini} introduces dynamic research blueprints and interactive plan refinement, leveraging Gemini 2.0 Flash Thinking's self-reflection capabilities and 1-million-token context window. 
Claude Research~\cite{claude2025research} adopts an agentic approach to conduct multiple mutually-building searches, automatically exploring different angles of a question. 
Perplexity Deep Research~\cite{perplexity2025deepresearch} integrates Bing-style indexing with the Sonar API, combining BM25 and dense vector reranking. 
On the open-source framework front, LangChain Open Deep Research~\cite{langchain2025open} provides a plan-and-execute architecture that identifies knowledge gaps through self-reflection. 
DeepResearchAgent~\cite{skywork2025deepresearch} employs a hierarchical multi-agent system. 
Together AI Open Deep Research~\cite{together2025open} generates search queries through initial planning and leverages LLMs to assess knowledge gaps. 
However, such systems essentially hardcode predefined workflow patterns into the system architecture, which imposes high requirements on system complexity. 
For agents implemented using approaches like ReAct, the research depth in specialized scenarios is often insufficient, failing to meet users' actual needs.

\subsection{End-to-End Optimization of Research Capabilities}
Unlike orchestration-based systems, some works internalize relevant capabilities into models through end-to-end training. 
DeepResearcher~\cite{zheng2025deepresearcher} conducts end-to-end RL training via GRPO in real web environments, demonstrating emergent cognitive behaviors such as planning, cross-source verification, and self-reflection. 
Kimi-Researcher~\cite{kimi2025researcher} supports long-horizon multi-turn search reasoning through end-to-end agentic RL training, employing context management mechanisms and asynchronous rollout systems. 
Tongyi DeepResearch~\cite{tongyi2025deepresearch} proposes a unified training paradigm for agentic mid-training and post-training, combining automated data synthesis pipelines with on-policy RL. 
Related agent training methodology works such as WebRL~\cite{webrl2025}, which proposes autonomous curriculum learning and KL-constrained policy updates, and Search-R1~\cite{searchr1}, which adopts RL-enhanced search reasoning integration, have made progress in this area.

Nevertheless, existing end-to-end works still primarily focus on improving \enquote{search efficacy} (e.g., retrieval accuracy, query optimization), lacking systematic construction strategies for \enquote{atomic core capabilities} crucial to in-depth research, such as long-horizon logical reasoning, multi-source cross-validation, and high-quality report composition. 
Furthermore, how to optimize model size and inference cost while maintaining high performance remains an urgent challenge for the industry. 

\subsection{Deep Research Related Evaluation Benchmarks}
In terms of evaluation, \textsc{ResearchRubrics}~\cite{sharma2025researchrubrics} contains 101 domain-diverse research tasks, each equipped with 20-43 expert-written fine-grained scoring criteria, assessing factual accuracy, reasoning soundness, and clarity. 
DeepResearch Bench~\cite{du2025deepresearch} comprises 100 PhD-level research tasks spanning 22 domains, evaluating report quality through the RACE framework and information retrieval capabilities through the FACT framework. 
ReportBench~\cite{li2025reportbench} reverse-engineers research questions based on expert survey papers on arXiv, assessing the citation accuracy and factual consistency of generated reports. 
However, these academic datasets for Deep Research, except for \textsc{ResearchRubrics}, remain insufficient in terms of task coverage comprehensiveness and evaluation depth, failing to adequately cover the diverse requirements of real research scenarios.

Another category of benchmarks focuses more on retrieval and knowledge testing capabilities. 
BrowseComp~\cite{wei2025browsecomp} contains 1,266 fact-finding questions requiring multi-hop reasoning, directly targeting the evaluation of models' ability to retrieve deeply hidden information. 
HLE (Humanity's Last Exam)~\cite{hendrycks2025hle} includes 2,500 expert-level questions across multiple disciplines, yet still carries a strong \enquote{closed-book exam} flavor, largely focusing on multi-hop search with definitive answers or frontier knowledge testing, failing to fully reach the openness and multidimensionality of real industrial-grade research scenarios. 
To this end, we construct ADR-Bench, aiming to fill the evaluation gap for Chinese Deep Research scenarios driven by real user demands.

\section{Data Strategy: Constructing Atomic Capabilities}
The core challenge of the Deep Research agent lies in bridging the decision-making gap between pre-training and task-specific optimization. During the pre-training phase, the model acquires vast world knowledge and language distributions; however, the post-training phase requires guiding the model to perform complex, long-horizon reasoning within a massive action space.

Direct exploration within the raw token-level space $\mathcal{A}_{\text{token}}$ is not only computationally expensive but also prone to trapping the model in local optima due to the exponential growth of the branching factor with sequence length. Therefore, we propose a unified data construction perspective: reshaping the training objective from ``predicting the next token'' to ``deciding the next \textbf{Atomic Action}.''

We define ``Atomic Capabilities'' as a set of transferable, high-level action abstractions that form a compact action subspace $\mathcal{A}_{\text{atomic}} \subset \mathcal{A}_{\text{token}}$. We formalize the goal of data construction as finding an optimal action subspace that minimizes two types of errors simultaneously:
$$\min (\epsilon_{\text{pruning}} + \epsilon_{\text{RL}})$$
where $\epsilon_{\text{pruning}}$ represents the approximation error caused by pruning potential optimal solutions, and $\epsilon_{\text{RL}}$ represents the difficulty of performing subsequent planning or reinforcement learning within this subspace.

To achieve Pareto optimality between ``retention of key skills'' and ``clarity of planning logic,'' we do not construct datasets in isolation. Instead, we establish query synthesis pipelines and trajectory generation strategies specifically around four core atomic capabilities: Planning \& Task Decomposition, Deep Information Seeking, Reflection \& Verification, and Reporting.


\subsection{Capability I: Planning \& Task Decomposition}

Planning capability requires the model to effectively decompose ambiguous or broad user requests into executable sub-tasks and to dynamically adjust its route based on environmental feedback. To achieve this, we adopt a ``Reverse Engineering'' strategy, utilizing existing ``perfect planning results'' from the real world to synthesize high-complexity planning data. Furthermore, to prevent tasks from being too trivial, we implement a rigorous screening process for task queries.

\subsubsection{Reverse Engineering Synthesis}
To obtain planning data that covers multiple domains and possesses authentic logical depth, we leverage high-quality documents such as open access technical reports, academic surveys, and financial research reports. These documents are essentially the final output of complex research tasks and contain implicit planning logic.

Specifically, we first take the title and abstract (or the original text) of a report and remove specific experimental details and result data. We then prompt an LLM to reverse-engineer the initial ``Project Task'' that could have led to this report, thereby generating high-difficulty queries that simulate real-world scientific and business scenarios. Simultaneously, utilizing the abstract structure as a form of ``hindsight,'' we synthesize a high-level Plan that guides the entire research process. This ensures that the generated planning path possesses extremely high feasibility and logic, serving as a strong constraint for the model during the inference phase.

Finally, to ensure data quality, we apply trajectory consistency filtering to the generated data. We generate execution trajectories for the agent and calculate their alignment with the preset plan, filtering out trajectories that complete the task but deviate significantly from the preset plan. This ensures the model learns an execution process that strictly conforms to the known hindsight.


\subsection{Capability II: Deep Search \& Information Seeking}

Deep information seeking capability differs from simple QA; it requires the model to possess capabilities for multi-hop reasoning, mining hidden entities, and active topological walking when information is incomplete. We specifically strengthen this capability through graph-based and multi-document synthesis pipelines.

\subsubsection{Graph-based Synthesis}
To construct questions requiring complex reasoning paths, we performed controlled subgraph sampling on open-source knowledge graphs such as Wikidata5m~\cite{wang2021KEPLER} and CN-DBpedia~\cite{2017CN}.

We adopt a specific topology construction strategy. First, we screen non-generic entities with small degrees (3-10) as ``seed nodes'' to avoid starting points that are too isolated or too broad. Subsequently, we perform BFS expansion centered on the seed to build a subgraph containing 10-40 nodes. To prevent semantic drift, we enforce truncation on super nodes with degrees exceeding a threshold (e.g., 1000), treating them as leaf nodes.

Given that triplets in knowledge graphs are often lossy and sometimes out-dated, we do not generate questions directly using triplets. Instead, for every edge in the subgraph, we use the triplet as a query to perform an additional search, verifying and faithfully expanding the triplet's information. Finally, based on this verified and structurally sound subgraph, we prompt an LLM to generate a fuzzed complex question requiring multi-hop search and reasoning, along with its corresponding answer.

\subsubsection{Multi-document-based Synthesis}
Addressing the capability of associative retrieval between documents, we utilized a custom index library, \textit{Wiki-doc}. Leveraging its natural hyperlink structure, we start from a random entity and use a few-shot prompt to guide a web search agent to perform a topology walk within the document index.

The agent is required to mine information by following hyperlinks without prior knowledge of the target, continuing until a maximum number of steps is reached or sufficient information is collected. Finally, we consolidate all node information collected along the path to reverse-generate $\langle \text{Query}, \text{Answer} \rangle$ pairs.

\subsubsection{Difficulty Filtering}
To ensure the non-triviality of planning tasks, we use QwQ-32b~\cite{qwq32b} as a difficulty filtering model. Any query that the QwQ-32b model can solve under the default ReAct framework is considered a ``simple problem'' and is excluded from the training set. Since QwQ-32b shares the same base model and pre-training knowledge as the model used in this work, while lacks extensive agent-specific training, tasks solvable by QwQ-32b can be regarded as simple tasks that do not require specialized training.


\subsection{Capability III: Reflection, Verification \& Cross-Validation}

In long-horizon reasoning, the model must possess the ability to identify its own errors (Self-Correction) and distinguish the authenticity of internet information (Fact-Checking). We designed specialized closed-loop pipelines to produce such data.

\subsubsection{The Error-Reflection Loop}
For deep information seeking queries, we employ a closed-loop process of ``expert model generation $\rightarrow$ result verification $\rightarrow$ multi-turn reflection'' to produce high-quality thought trajectories. This process aims to improve the model's anti-interference ability and cross-validation level in complex retrieval environments through introspection on failed paths.

The specific synthesis pipeline is as follows. An expert-level model generates a preliminary search and reasoning trajectory. If the final output matches the answer, the trajectory is included directly in the training set as a positive sample. If the result does not match, we construct a prompt asking the model to perform a reflection action based on the incorrect result. Based on the reflection conclusion, the model retains historical memory and retries the trajectory generation; this process iterates up to 3 times.

For trajectories that eventually reach the correct answer after reflection, we perform specialized post-processing cleaning. We remove phrases with traces of artificial induction, such as ``according to user hints,'' ensuring the output appears as the model's completely spontaneous introspection and error-correction process. This strategy not only strengthens the ability to gather unknown information but also teaches the model to filter unreliable internet data by comparing multiple sources, significantly reducing factual bias in the final report.

\subsubsection{Deep Verification Workflow}
To strengthen the factual rigor of the model in complex research tasks, we cleaned thousands of $\langle \text{paragraph}, \text{judge-result} \rangle$ pairs from desensitized real data as seed samples. We constructed a multi-agent teacher workflow to simulate the verification process of human experts, recording the complete execution path as an agent trace. The system consists of the following collaborative atomic agents:

\begin{itemize}
    \item \textbf{Extract Agent:} Analyzes input materials and performs factual decomposition. It extracts time, location, subjects, core data, and causal events from natural language descriptions, converting them into independent verification points.
    \item \textbf{Plan Agent:} Generates a preliminary action plan. It analyzes the necessity of verification and the required source type for each verification point produced by the extract agent, generating a plan based on logical dependencies.
    \item \textbf{Verify Agent:} Executes specific verification actions. It calls search tool and models to perform multi-source retrieval and cross-validation on the content.
    \item \textbf{Replan Agent:} Executes dynamic path adjustment. It summarizes currently acquired information and adjusts the research direction in a timely manner, significantly reducing redundant searches and improving decision quality under complex paths.
    \item \textbf{Report Agent:} Aggregates evidence from the entire process. It provides a clear final conclusion (support, refute, or doubtful) for each verification point, accompanied by complete citation evidence.
\end{itemize}

The generated verification trajectories undergo strict posterior filtering. The verification point, conclusion, and evidence are treated as a triplet and input into a judge model to verify whether the report conclusion is logically self-consistent with the evidence, ensuring that the model learns a verification paradigm that is both factually accurate and logically rigorous.


\subsection{Capability IV: Report Generation}

Report writing is not merely text generation, it is a process of structured reorganization of collected fragmented information. We divide the training of this capability into a \textit{mid-training} phase, focusing on domain style and content depth, and an \textit{SFT} phase, focusing on instruction following and formatting specifications.

\subsubsection{Mid-training: Domain Style and Content Depth}
In the mid-training phase, the goal is for the model to internalize expert-level writing frameworks and terminological styles. We constructed large-scale $\langle \text{Query}, \text{Report} \rangle$ pairs. The data originates from strictly screened high-quality human reports (such as financial research reports and in-depth surveys), while the queries are obtained through the aforementioned reverse engineering method. In this stage, the model focuses on learning how to organize language, cite data, and develop in-depth arguments like an expert within the context of a given project task, without focusing on the specific search process.

\subsubsection{SFT: Instruction Following and Formatting Specifications}
In the SFT phase, the focus shifts to instruction following regarding user-specific constraints and planning consistency. For Deep Research queries with metadata containing a plan, we require the report generated by the model to strictly follow the preset plan structure. We generate trajectories and perform alignment checks with the plan, filtering out samples that deviate. For samples with high trajectory generation quality but poor report formatting, we employ a specialized System Prompt to regenerate the report based on the final round state, ensuring the final output is not only detailed in content but also precisely responsive to user needs in terms of instruction following and formatting.

\section{Training Pipeline}
A progressive three-stage training pipeline is adopted on top of a 32-billion-parameter base model, consisting of mid-training, supervised fine-tuning (SFT), and reinforcement learning (RL), which are applied sequentially to systematically enhance the model's overall performance in complex reasoning and long-horizon agent tasks. A clear division of responsibilities across capability expansion, behavior alignment, and objective optimization is established by this training design. In particular, high-quality atomic capability data are introduced during the mid-training stage, including planning and task decomposition ability, deep search and information seeking ability, reflection and verification ability, and report generation ability. In the subsequent SFT stage, these atomic capabilities are composed, while explicit constraints and alignment are imposed on instruction following, output structure, and task-specific formatting. Finally, task-level feedback signals are incorporated in the RL stage, through which the behavioral quality of the model in realistic interactive and decision-making scenarios is further optimized.

Qwen2.5-32B-Base~\cite{qwen2024tech} is selected as the base model to achieve a balance among performance, computational cost, and experimental reproducibility. Through this choice, near–large-scale core capabilities and long-context support are retained, while the barrier to multi-stage training and systematic ablation studies is substantially reduced. Consequently, the performance improvements observed in this work can be more directly attributed to the proposed training paradigm and data strategies, ensuring clearer interpretability and reproducibility. As a medium-scale model, core capabilities comparable to those of 72B models such as strong instruction following and logical reasoning are exhibited by Qwen2.5-32B-Base, together with support for up to 128k context length, making it well suited for long-horizon agent tasks. Meanwhile, the 32B-parameter scale significantly lowers training costs, and strong model plasticity has been demonstrated in prior fine-tuning efforts. By using Qwen2.5-32B-Base as the base model, it is ensured that the performance gains reported in Deep Research scenarios are objectively derived from the data strategies and training paradigm, thereby facilitating community reproduction and fair comparison.

\subsection{Stage 1: Agentic Mid-training}
To systematically enhance the model’s capabilities in long-context understanding, knowledge integration, and tool-augmented reasoning for complex tasks without significantly increasing the parameter scale, we introduce a two-stage mid-training process between pre-training and instruction fine-tuning. This stage is designed to progressively adapt the model to more complex and longer-sequence tasks through carefully constructed data distributions and context-length scheduling, enabling the emergence of stable medium to long-horizon reasoning as well as task decomposition and execution capabilities.

Overall, mid-training follows a curriculum that progresses from shorter to longer contexts and from pure knowledge-based tasks to tool-augmented tasks, and is divided into two stages: the first stage supports a maximum context length of 32K, while the second stage extends this capacity to 128K.

\paragraph{Mid-training (32K Context).} The first-stage mid-training focuses on injecting atomic capabilities such as planning and task decomposition, information seeking, reflection and verification, and report generation. This stage aims to substantially enhance the model’s ability to understand and execute complex task structures under medium-length contexts. Rather than directly optimizing for task completion rates, this phase emphasizes the systematic construction of high-quality capability-oriented data, enabling the model to form stable and generalizable intermediate representations across key dimensions including planning, evidence integration, and self-correction. These representations serve as a solid foundation for subsequent training stages involving longer contexts and real-world interactive agent tasks. The training data in this stage primarily covers capability dimensions such as planning, information seeking, reflection, and report generation. The detailed composition is shown in Table~\ref{midtrain-stage1}.
\begin{table}[!ht]
\centering
\caption{Data Composition of Mid-training (32K Context)}
\label{midtrain-stage1}
\begin{adjustbox}{width=0.98\linewidth}
\begin{tabular}{l l l l}
\hline
\textbf{Data Category} & \textbf{Data Source} & \textbf{Primary Format} & \textbf{Capability Focus} \\
\hline
Active reading general knowledge data & Wiki, Baidu Baike & QA, Rewrite & Information seeking \\
Active reading academic data & Academic articles & QA, Rewrite & Information seeking \\
Synthetic knowledge data & PleIAs SYNTH\footnotemark[1] & Synthetic text, structured QA, reasoning traces, multilingual & Report generation + information seeking \\
Summarization data & In-house / mixed & Document $\rightarrow$ summary & Report generation \\
Reasoning data & In-house / mixed & Multi-step reasoning text & Planning and task decomposition \\
Reflection data & In-house / mixed & Self-checking and correction & Reflection and verification \\
\hline
\end{tabular}
\end{adjustbox}
\end{table}

\footnotetext[1]{\url{https://huggingface.co/datasets/PleIAs/SYNTH}}

Among these, the academic active reading data simulates the model’s realistic cognitive process when reading long-form academic documents, guiding it to learn how to identify critical information from large volumes of text and to form effective internal knowledge representations through question-answering and rewriting tasks. Explicit tool invocation is not introduced at this stage, ensuring that the model develops robust comprehension and reasoning capabilities under pure textual conditions.

\begin{figure}[!ht]
    \centering
    \includegraphics[width=0.7\linewidth]{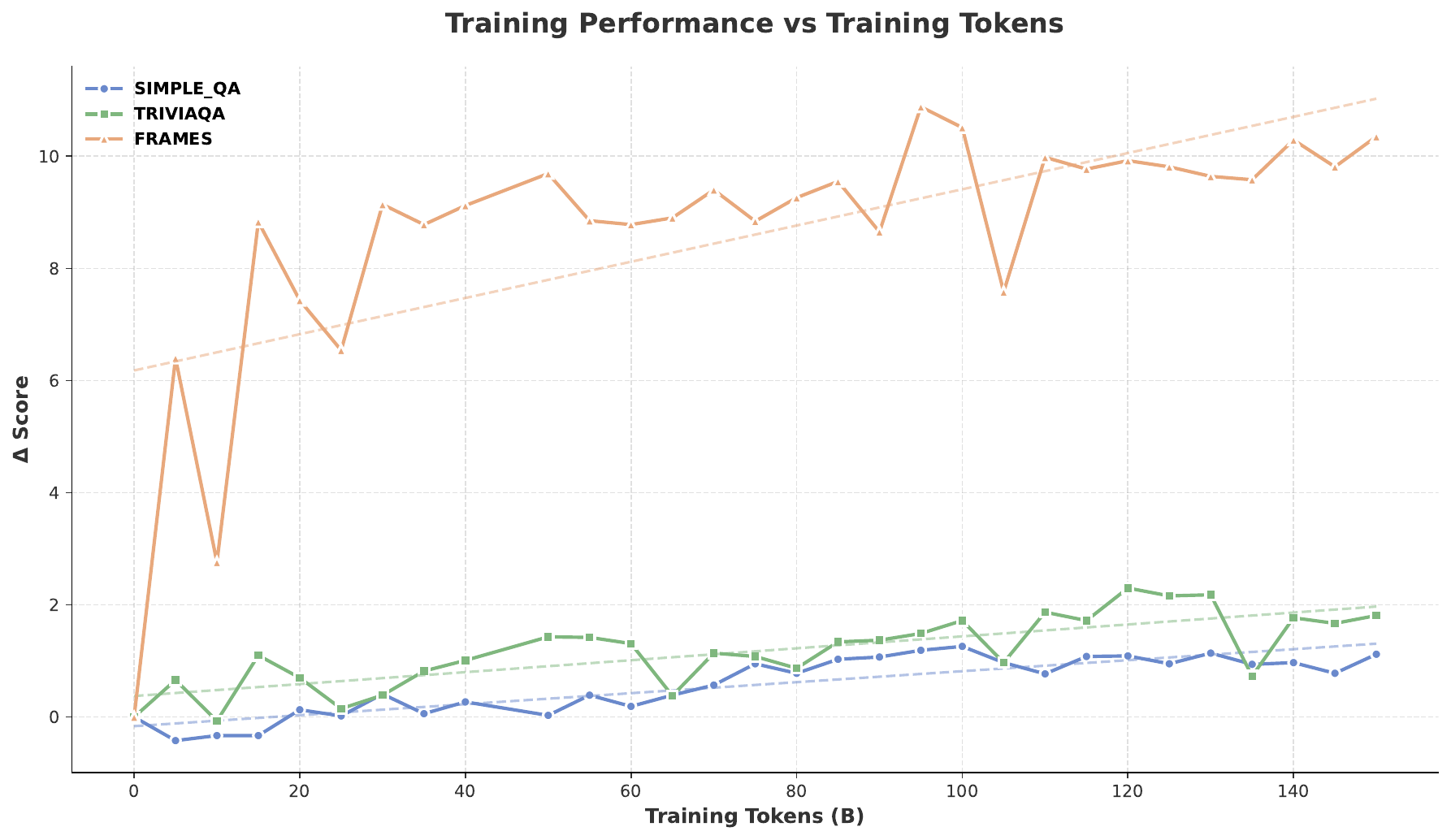}
    \caption{
    \textbf{Performance trends during mid-training with 32K context.} Average accuracy on SimpleQA, TriviaQA, and FRAMES is reported at checkpoints saved every ~5B tokens. Performance improves steadily with training token scale, with particularly large gains on FRAMES, indicating strengthened agentic and structured reasoning. Curves remain unconverged at 150B tokens, suggesting further headroom.
    }
    \label{fig:midtrain-performance-stage1}
\end{figure}

As shown in Figure \ref{fig:midtrain-performance-stage1}, during training, checkpoints are saved approximately every 5 billion tokens, and performance gains are analyzed over the full 150-billion-token mid-training process in Stage 1. The results indicate that as the training data scale increases, the model exhibits consistent performance improvements on the SimpleQA, TriviaQA, and FRAMES benchmarks. The maximum observed gains reach approximately +1.26\% on SimpleQA, +2.30\% on TriviaQA, and +10.88\% on FRAMES. Overall, the 32K mid-training stage has not yet fully converged at the 150-billion-token scale, suggesting substantial room for further improvement, particularly in agent-related and structured reasoning capabilities.

\paragraph{Mid-training (128K Context).} Building upon the stable convergence of capabilities achieved in Stage 1, the second stage of mid-training further extends the maximum context length to 128K, with a primary focus on real-world complex task scenarios. This stage strengthens the model’s abilities in retrieval, planning, and tool-augmented reasoning under ultra-long contexts. The training data in this stage is more closely aligned with practical applications and covers complex structures such as web interaction, search processes, and multi-tool collaboration. The detailed data composition is summarized in Table~\ref{table:Stage2-Mid-training}.

\begin{table}[!h]
\centering
\caption{Data Composition of Mid-training (128K Context)}
\label{table:Stage2-Mid-training}
\begin{adjustbox}{width=0.98\linewidth}
\begin{tabular}{l l l l}
\hline
\textbf{Data Category} & \textbf{Data Source} & \textbf{Primary Format} & \textbf{Capability Focus} \\
\hline

URL QA data & In-house / mixed & URL + QA & Information seeking + planning and task decomposition \\
Deep Search data & In-house & Search steps + tool calls & Information seeking + planning and task decomposition \\
Web browsing data & Agent-data-collection~\cite{song2025agent} & Web navigation + tool calls & Information seeking + planning and task decomposition \\
Planning data & In-house / mixed & Task planning + tool calls & Planning and task decomposition \\
Summarization data & In-house / mixed & Long document / long dialogue summarization & Report generation \\
Reasoning data & In-house / mixed & Long-horizon dialogue reasoning & Planning and task decomposition \\
Reflection data & In-house / mixed & Reflection and correction processes & Reflection and verification \\
General dialogue data & High-quality dialogue corpora & Multi-turn dialogue & Maintaining general language interaction ability \\
\hline
\end{tabular}
\end{adjustbox}
\end{table}

By introducing explicit tool invocation at this stage, the model is required not only to generate coherent intermediate reasoning processes, but also to learn when and how to select and invoke external tools, and to integrate the returned results into subsequent reasoning. This significantly enhances the model’s practical effectiveness in agent-style tasks.

To more clearly illustrate the design differences and the progressive relationship between the two mid-training stages, key dimensions are compared in Table \ref{table:compare-two-stages-midtrain}.

\begin{table}[t]
\centering
\caption{Comparison of the Two Mid-training Stages}
\label{table:compare-two-stages-midtrain}
\begin{adjustbox}{width=0.98\linewidth}
\begin{tabular}{l c c}
\hline
\textbf{Dimension} & \textbf{Mid-training Stage I} & \textbf{Mid-training Stage II} \\
\hline
Maximum context length & 32K & 128K \\
Primary objective & Knowledge injection/Basic comprehension & Retrieval/Planning/Tool-augmented reasoning \\
Data focus & Encyclopedic/Academic/Cognitive patterns & Web/Search/Tool invocation \\
Tool calling included & No & Yes \\
Task complexity & Medium & High \\
Application scenarios & Understanding/Reasoning/Simple QA & Deep search/Agent scenarios \\
\hline
\end{tabular}
\end{adjustbox}
\end{table}

\subsection{Stage 2: Post-training Supervised Fine-tuning}
During the mid-training stage, the model is equipped with fundamental atomic capabilities such as planning and information seeking. In the post-training supervised fine-tuning (SFT) stage, the focus shifts away from isolated capability teaching toward the composition of these atomic abilities to improve end-to-end performance on long-horizon tasks. The core objective of this stage is domain adaptation and performance enhancement: by leveraging rigorously cleaned, high-quality full-chain trajectories, the model’s existing atomic capabilities are systematically connected to form efficient, robust, and behavior patterns deeply aligned with the requirements of Deep Research scenarios.

\paragraph{SFT Data Composition.}
The SFT dataset primarily consists of two categories of end-to-end task trajectories: \emph{Deep Search} and \emph{Deep Research}.

\begin{itemize}
    \item \textbf{Deep Search data.}  
    This category focuses on multi-hop search tasks with well-defined ground-truth answers. Although a portion of deep search trajectories is already introduced during the mid-training stage to establish basic retrieval and reasoning capabilities, the SFT stage emphasizes performance optimization and stylistic diversity. High-quality trajectories with uniformly distributed queries are selected, and diverse reasoning patterns are introduced to enable the model to flexibly choose optimal reasoning paths for different queries. This results in a qualitative transition from merely being able to retrieve correct information to doing so both efficiently and accurately.
    
    \item \textbf{Deep Research data.}  
    This category targets comprehensive research tasks involving open-ended questions. The data covers the full pipeline of intent understanding, planning, information cross-verification, and report generation under strict formatting requirements (e.g., citation and attribution). Through this design, the entire end-to-end logic of \emph{intent-analysis–planning–execution–reflection–writing} is reinforced.
\end{itemize}

\paragraph{Data Construction and Cleaning Strategies.}
The effectiveness of SFT is highly dependent on data quality. To this end, a refined data pipeline is constructed with strict rules and algorithmic filtering to ensure high-quality supervision. Representative strategies include:

\begin{itemize}
    \item \textbf{Trajectory efficiency optimization.}  
    For deep search tasks with ground-truth answers, a ``correct and shortest'' principle is applied during data cleaning. Among all successful trajectories, only those with the fewest reasoning steps and the most concise tool usage are retained. This encourages the model to replace redundant search behaviors with effective reasoning and to acquire information at minimal cost, thereby eliminating unnecessary tool invocations.
    
    \item \textbf{Robustness and noise control.}  
    A controlled proportion of trajectories containing tool-call errors is intentionally retained, such as empty search results or tool failures followed by correct \emph{reflection–correction} actions. This form of structured noise injection prevents the model from collapsing under real-world web instability and equips it with self-correction capabilities for handling exceptional cases.
    
    \item \textbf{Cognitive pattern deduplication.}  
    A strict $N$-gram–based deduplication mechanism is applied to identify and remove low-quality trajectories with excessive repetition or degenerate loops. This ensures diversity and flexibility in the model’s reasoning behavior during long-horizon tool usage.
    
    \item \textbf{Strict citation and factual alignment.}  
    To meet the rigor required for Deep Research reports, citation formats using \verb|\cite{}| are explicitly incorporated into the SFT data. This construction guides the model to adopt a writing paradigm in which relevant references are appended at critical information points, ensuring traceability and factual grounding, and aligning the output format with professional research standards.
\end{itemize}

\subsection{Stage 3: Reinforcement Learning}
In the first two stages, the training process primarily relies on large-scale synthetic and distilled data for Mid-training and Supervised Fine-tuning (SFT). Although this approach is effective for instruction-following tasks and basic tool usage, it mainly depends on ``teacher'' trajectories and cannot be improved through trial-and-error learning with real-world interactions. Therefore, we introduce reinforcement learning (RL), connecting the model directly to the real tool usage environment and optimizing it through interactions with the environment. Unlike short-term searches focused on entity matching, the quality of Deep Research reports depends on multiple dimensions, including task decomposition, planning, tool invocation strategy, evidence selection and validation, and the final report generation. 
A recent approach is to transform report quality evaluation into a set of rubric-based scores, which are then used as optimization signals~\cite{gunjal2025rubrics}. This rubric-based reinforcement learning enhances the model's abilities in active planning, reflection, and cross-source validation, overcoming the limitations of offline imitation and significantly improving both performance and user experience.

\subsubsection{RL Data Synthesis}
Although small-scale benchmarking can rely on expert-defined rubrics~\cite{sharma2025researchrubrics}, the reinforcement learning training still requires a data synthesis process due to the difficulty and high cost of collecting high-quality rubrics. 

\paragraph{Two-Step Reverse Synthesis.}
To generate Deep Research tasks and their corresponding rubrics, we employ a ``two-step reverse synthesis'' approach. Traditional forward extraction methods typically derive rubrics by rephrasing or decomposing user queries. However, these methods often depend on surface-level expressions and may overlook key quality dimensions implicit in the task. Therefore, we designed a more refined synthesis process:
\begin{itemize}
    \item In the first step, guided by a small number of high-quality examples and templates, we use a powerful LLM to generate an initial task description (referred to as the hidden task summary) and concurrently generate a set of fine-grained rubrics. Each rubric is defined as an atomic standard containing evaluation dimensions and importance weights, ensuring that the standard is specific, verifiable, and practically applicable in task assessment. Furthermore, each rubric is assigned a role indicating whether it is an explicit, implicit, or negative requirement, corresponding to whether it needs to be explicitly stated in the task or serves as a bonus or penalty item. 
    \item In the second step, we synthesize the target task (i.e., the actual user task query) based on the synthesized rubrics. During this process, we simultaneously re-assess the role of each rubric for the synthesized task and provide brief reasoning and attribution. If any discrepancies are found between the initial and final role assignment, the entire synthesis sample is discarded. This ensures that the synthesized task queries align with the rubric's requirements, semantically support the key dimensions in the rubric, and cover both explicit and implicit task requirements.
\end{itemize}

\paragraph{Objective Consistency Verification.}
To ensure that the objectives and requirements of the task remain aligned during the reverse synthesis process, we implement an objective consistency verification step to check the consistency between the hidden task summary, rubrics, and the synthesized task. This process is performed by an independent judge model, which evaluates the consistency between the hidden task summary and the synthesized task, and the alignment of the rubrics with the task requirements. The judge model assesses the relevance of each rubric and ensures no contradictions with the task objectives. 
Finally, it provides an overall consistency score and decides whether the sample should be retained. This step ensures high-quality task samples and reliable supervision signals for subsequent reinforcement learning training.

\subsubsection{Reward Design}
\paragraph{Rubrics Judge Training.}
In the Deep Research report scenario, relying directly on large LLMs to assign fine-grained scores based on the full rubrics for each report would require dozens of inferences per sample to cover all evaluation dimensions. This approach incurs prohibitively high computational costs and latency during large-scale agentic RL training. Based on our observation, the quality of the rubrics is often more critical in ensuring that the LLM Judge's annotations align with human experts, rather than the absolute performance of the judge model itself.

Thus, we first use a strong model to generate scores and explanations on the constructed $\langle\text{Query, Rubrics, Report}\rangle$ triples, which then serve as supervisory signals for training the rubrics judge model. This model undergoes two training phases: supervised fine-tuning and reinforcement learning with verifiable rewards. The former aims to help the model learn the scoring logic and explanation style of the strong model, establishing foundational discriminative ability, while the latter further strengthens the consistency of the model output format and the robustness of the scoring logic. Ultimately, this Rubrics Judge serves as the primary reward provider in this stage, allowing us to support large-scale agentic RL training within an acceptable budget without compromising signal quality.

\paragraph{Strict Reward Mapping.}
In our initial attempt to evaluate reports based on rubrics, we categorized each rubric's judgment into three classes: ``fully satisfied'', ``partially satisfied'' and ``not satisfied'', mapping them directly to ternary judgments $\{1, 0.5, 0\}$. However, we found that the agreement between different strong models (acting as LLM judges) and between LLM judges and human experts was lower in the ``partially satisfied'' category. This misalignment in the intermediate category may lead to unstable reward signals, making it difficult to provide consistent optimization directions and weakening the model's ability to adhere to constraints.
To ensure more distinguishable reward signals, we convert the ternary judgments into binary signals based on the nature of the rubrics: 
\begin{itemize}
    \item \textbf{For positive rubrics:} We map ``fully satisfied'' to $1$ and unify ``partially satisfied'' and ``not satisfied'' as $0$. This approach emphasizes the principle of ``no reward unless fully satisfied'', preventing the model from exploiting low-quality generations.
    \item \textbf{For negative rubrics:} We map ``not satisfied'' to $0$, while ``partially satisfied'' and ``fully satisfied'' are unified as $1$. This ensures that any deviations from the desired outcome are penalized, helping the model avoid undesirable behaviors.
\end{itemize}
The report's final reward will consider the judgment of each rubric and its corresponding weight (which can be positive or negative). This asymmetric binary mapping eliminates noise from intermediate categories, making the reward signals more discriminative and accelerating the convergence of the model toward expert-aligned behavior.

\subsubsection{Agentic Reinforcement Learning}
\paragraph{Real-world Environment with Tools.}
To enable the agent to genuinely learn how to complete deep-research tasks under real-world constraints, we place it in a multi-tool interactive environment during the reinforcement learning stage. Within this environment, the agent can freely alternate between natural-language generation and tool invocation. For cost control, we introduce a caching mechanism for high-expense web retrieval, reusing results for identical queries, and impose explicit budgets on the number of tool calls and total token consumption to ensure that training operates within a predictable resource envelope.
Within this environment, each deep-research task is treated as a reinforcement learning episode. The agent first performs intent analysis and restates the user query, then produces a step-by-step plan, organizing tool invocations and information gathering around distinct subproblems. It subsequently filters, contrasts, and cross-validates evidence from multiple sources, and ultimately drafts a complete report. After the episode terminates, the Rubrics Judge evaluates the report against the predefined rubrics, producing multi-dimensional scores that are converted—via the strict reward mapping into reward signals used for policy updates. By iterating this process, the model continuously learns through trial-and-error in a real environment, progressively acquiring policy decisions that more closely match human expert preferences with respect to tool selection, invocation timing, and call ordering.

\paragraph{PPO Algorithm.}
We model the deep-research agent’s behavior in a multi-tool environment as a sequential decision-making process governed by a single policy $\pi_\theta$. At each time step $t$, given a state $s_t$—comprising the user request, the history of generated tokens, and observations returned by tools—the policy outputs an action $a_t$. Here, an action includes both natural-language token generation and structured decisions such as initiating tool calls, specifying tool parameters, and parsing tool outputs. A complete interaction yields a trajectory $ \tau=\{{(s_t,a_t)}\}_{t=1}^{T} $, and, upon episode termination, a terminal reward $ R(\tau) $ is assigned by the Rubrics Judge. Under this formulation, we perform on-policy optimization using the clipped Proximal Policy Optimization (PPO) objective:
\begin{equation}
\max_{\theta} \; \mathcal{J}(\theta)
= \mathbb{E}_{\tau \sim \pi_{\theta_{\text{old}}}}
\Bigg[
\sum_{t=1}^{T}
\min \Big(
r_t(\theta)\,\hat{A}_t,\;
\mathrm{clip}\big(r_t(\theta),\, 1-\epsilon,\, 1+\epsilon\big)\,\hat{A}_t
\Big)
\Bigg].
\label{eq:ppo-objective}
\end{equation}

Among these, the importance ratio is
\begin{equation}
r_t(\theta)=\frac{\pi_{\theta}(a_t\mid s_t)}{\pi_{\theta_{\text{old}}}(a_t\mid s_t)}.
\label{eq:importance-ratio}
\end{equation}
This objective employs $\epsilon$-clipping to explicitly bound the policy shift induced by each update, thereby maintaining training stability in settings characterized by long-horizon sequence generation and sparse terminal rewards. For advantage estimation, we use Generalized Advantage Estimation (GAE):
\begin{equation}
\hat A_t^{\text{GAE}(\gamma,\lambda)}=\sum_{l=0}^{T-t}(\gamma\lambda)^l\delta_{t+l},\qquad
\delta_t=r_t+\gamma V_{\phi}(s_{t+1})-V_{\phi}(s_t),
\label{eq:GAE}
\end{equation}
Consistent with the engineering practice in Open-Reasoner-Zero, we set $ \gamma=1 $ and $\lambda=1$ in GAE, i.e., we apply neither discounting to future returns nor additional exponential smoothing. This choice substantially simplifies and accelerates the computation of credit assignment in long-horizon sequences with sparse rewards.

We choose PPO over critic-free policy-gradient variants primarily because a learned critic can provide more reliable token-level value estimation. The systematic analysis in Open-Reasoner-Zero indicates that a critic can identify and down-weight harmful patterns such as repetitive loops, thereby yielding more robust advantage estimates and improving training stability. In contrast, methods without an explicit value function have greater difficulty distinguishing “accidentally high returns” from returns coupled with undesirable behaviors, which can lead to erroneous reinforcement and instability. This consideration is particularly important in agentic settings and deep-research tasks: report generation typically involves long horizons, cross-source retrieval, and multi-round tool use. If a single terminal, rubric-based reward cannot be assigned in a fine-grained manner to intermediate decisions (e.g., retrieval strategy, evidence selection, cross-validation, and structured writing), policy updates tend to collapse into a coarse “overall good/bad” fit. PPO’s actor–critic structure directly supports this need by providing actionable token-level learning signals.

\paragraph{Training dynamics.}
\begin{figure}[ht]
    \centering
\includegraphics[width=0.5\textwidth]{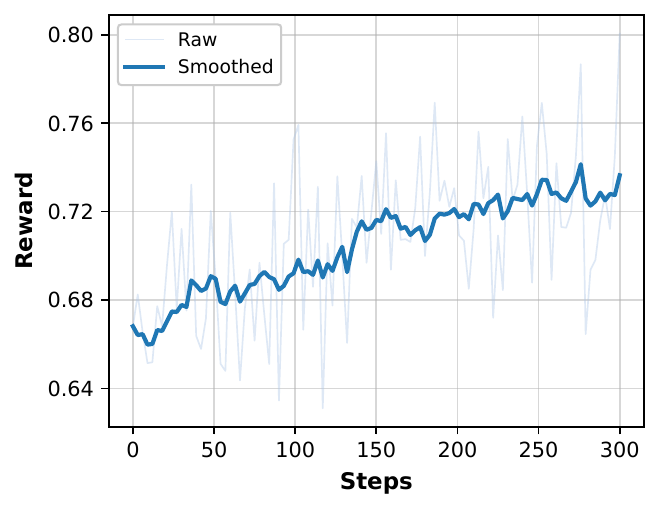}
    \vspace{-1em}
    \caption{{Training reward of RL}}
    \label{rl-reward}
\end{figure}

The training dynamics of the RL process are presented in Figure \ref{rl-reward}. The reward curve exhibits a consistent upward trajectory as training progresses, demonstrating that the agent is effectively optimizing its policy within the task distribution.

\section{System Architecture}
\begin{figure}[!ht]
    \centering
    \includegraphics[width=0.8\linewidth]{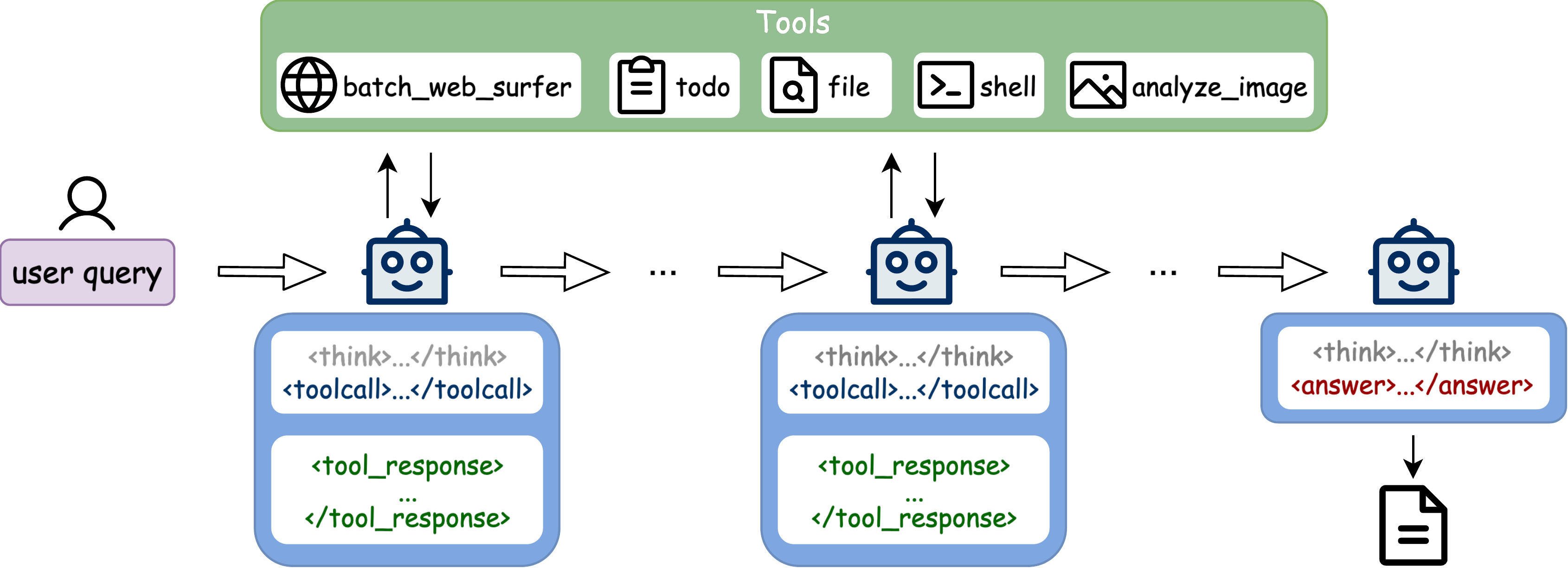}
    \caption{\textbf{Step-DeepResearch System Architecture.} The agent operates within a ReAct loop, utilizing a specialized toolset (e.g., \textit{batch\_web\_surfer}, \textit{todo}, \textit{shell}) for planning, execution, and reflection to generate comprehensive research reports.}
    \label{fig:framework}
\end{figure}

Step-DeepResearch is implemented within a basic single-agent architecture following the ReAct paradigm, reframing complex Deep Research tasks into a dynamic reasoning-action-observation loop. As shown in Figure~\ref{fig:framework}, upon receiving a user query, the agent initiates a cognitive iterative process through explicit reasoning and tool invocation, which cycles through three core phases.

\paragraph{Planning \& Reflection.} The agent initially identifies user intent to formulate action plans. In subsequent turns, it spontaneously reviews prior outcomes to validate the current state against objectives, achieving dynamic self-correction.

\paragraph{Tool Execution.} Translating plans into concrete actions, the agent selects the most appropriate tool set ( e.g., \textit{batch\_web\_surfer} for search, \textit{todo} for tracking ) to initiate precise data acquisition requests.

\paragraph{Feedback \& Cross-Validation.} New tool response is injected into the next reasoning round. The agent performs cross-validation against historical context—resolving conflicts and filtering falsehoods—to construct a logically rigorous chain of evidence.

This mechanism allows the agent to autonomously determine the exploration depth until the final report is generated. To support the characteristics of these Deep Research tasks, which involve long-horizon reasoning, strong interaction, and high uncertainty, we design a unified tool system. The system is built upon several guiding principles, including reusing human interaction logic, structuring feedback information, preserving original interaction styles, and unifying tool execution through a layered abstraction, interwoven with the agent's execution loop. The core design philosophy consists of aspects: \textbf{capability alignment}, providing operational capabilities comparable to those of humans when conducting research tasks on a computer; \textbf{information adaptation}, emphasizing structured and complete tool outputs with high information density, while explicitly accounting for context window constraints and failure recovery; and \textbf{architectural simplification}, merging functionally similar tools as much as possible while preserving flexibility and expressiveness. Detailed below are the core components of this infrastructure:

\paragraph{Authoritative Enhanced Information Acquisition.} To address the imbalance of signal-to-noise ratio and the dilution of authority in massive internet data, we built a high-quality information retrieval system.
\begin{itemize}
    \item \textbf{Curated Authority Indexing:} We assembled a professional team to evaluate diverse online sources, selecting over 600 core authoritative sites covering official government domains, industry research institutes, international organizations, and leading academic platforms. We built independent index shards for these sites, physically and logically isolating authoritative content from low-quality SEO spam, significantly improving recall stability.
    \item \textbf{Knowledge-Dense Document Retrieval:} Targeting long-form documents such as professional literature, financial reports, and official white papers, the system utilizes a dedicated library of over 20 million high-quality items. Indexing employs paragraph-level granularity to avoid the noise introduced by ingesting entire documents. By recalling specific semantic paragraphs, the model acquires higher-density information with lower token costs.
    \item \textbf{Authority-Aware Ranking Heuristics:} In the ranking phase, the system integrates an authority boosting factor. When semantic relevance scores are comparable, the algorithm prioritizes content from authoritative sites, ensuring that research arguments are grounded in verifiable facts. Additionally, customized search parameter interfaces allow the system to dynamically adjust search behavior based on query intent.
\end{itemize}

\paragraph{Knowledge Management \& File Operations.} Treating the file system as an external persistent memory, we adapted traditional file interactions into agent-native protocols. This transformation is strictly driven by the need for token efficiency and robustness in long-context workflows.
\begin{itemize}
    \item \textbf{Token-Efficient Patch-based Editing:} To mitigate the token waste of full rewrites and the reasoning burden imposed by diff formats on mid-sized models, we introduced a \textit{patch} action. The agent only needs to provide the modified fragment with minimal anchor context, while the tool performs atomic updates via fuzzy matching. In scenarios involving local polishing of long-form reports, this reduces output token costs by over 70\% and significantly boosts the success rate of complex edits.
    \item \textbf{Implicit Context Management:} To eliminate the risk of context window overflow caused by excessive retrieval content, we devised a summary aware local storage strategy. When tool results exceed a preset threshold, the system truncates the immediate return, injecting only high-density summaries into the context while persisting raw data to local temporary files. The agent performs demand-paging via \texttt{file.read} based on summary cues. This design effectively offloads context pressure to disk, enabling virtually infinite context support for long-horizon reasoning.
    \item \textbf{Stateful Todo Management:} To prevent goal drift in long-horizon research, the \texttt{todo} tool encapsulates complex CRUD operations within a unified entry point. It automatically determines \enquote{create, rewrite, or destroy} states based on the current task stack. By decoupling research progress from model weights and persisting it at the tool layer, this ensures logical consistency and goal alignment over extensive interaction trajectories.
\end{itemize}

\paragraph{Interactive Execution \& Multimodal Perception.} To bridge the gap between raw execution and expert-level problem solving, we developed a high-fidelity interaction framework integrated with a comprehensive multimodal perception suite. This setup ensures the agent can navigate complex digital tasks with both precision and adaptability.
\begin{itemize}
    \item \textbf{Human-like Terminal Interaction (Sandbox \& Tmux Integration):} All execution commands are run within a restricted MCP sandbox environment. The system integrates the \texttt{tmux} session management mechanism; by maintaining a persistent scrollback buffer, the agent can stably operate command-line programs with state-refreshing characteristics (such as the \texttt{vim} editor or real-time monitoring tools). This equips the agent with human-like interactive debugging capabilities when facing complex errors, enhancing system-level fault tolerance.
    \item \textbf{Perception-optimized Resilient Browser:} We introduce a visual redundancy elimination strategy specifically for research tasks. The system dynamically identifies visual differences by calculating the \textit{PHash} (Perceptual Hash) distance between page screenshots of consecutive actions. In cases of negligible page updates, the system inhibits image-based feedback and reverts to a text-only stream. This strategy ensures the agent maintains real-time control over page states while significantly reducing multimodal token redundancy.
    \item \textbf{Multimodal Perception Tools:} For unstructured data, we have integrated specialized modules including \texttt{file\_parser} (document parsing), \texttt{asr} (audio transcription), and \texttt{analyze\_image} (image analysis), ensuring high-quality understanding of complex research materials.
\end{itemize}

\section{ADR-Bench: A Custom Deep Research Benchmark}
As a product designed for complex, open-ended research tasks, Deep Research faces long-standing evaluation hurdles, characterized by inconsistent standards, inherent subjectivity, and high expertise thresholds. To steer model iterations scientifically and efficiently while reflecting authentic user perception, we implemented a dual-track evaluation framework covering both general-domain user experience and professional-domain capability assessments. This chapter systematically summarizes our core methodologies, key findings, and empirical experiences derived from our benchmark construction and evaluation design. It details query composition, evaluation framework architecture, and criteria formulation, alongside the challenges and exploratory initiatives undertaken, aiming to provide a reference for the evaluation of similarly complex tasks.

\begin{figure}[ht]
    \centering
\includegraphics[width=0.6\linewidth,
  height=0.4\textheight,
  keepaspectratio]{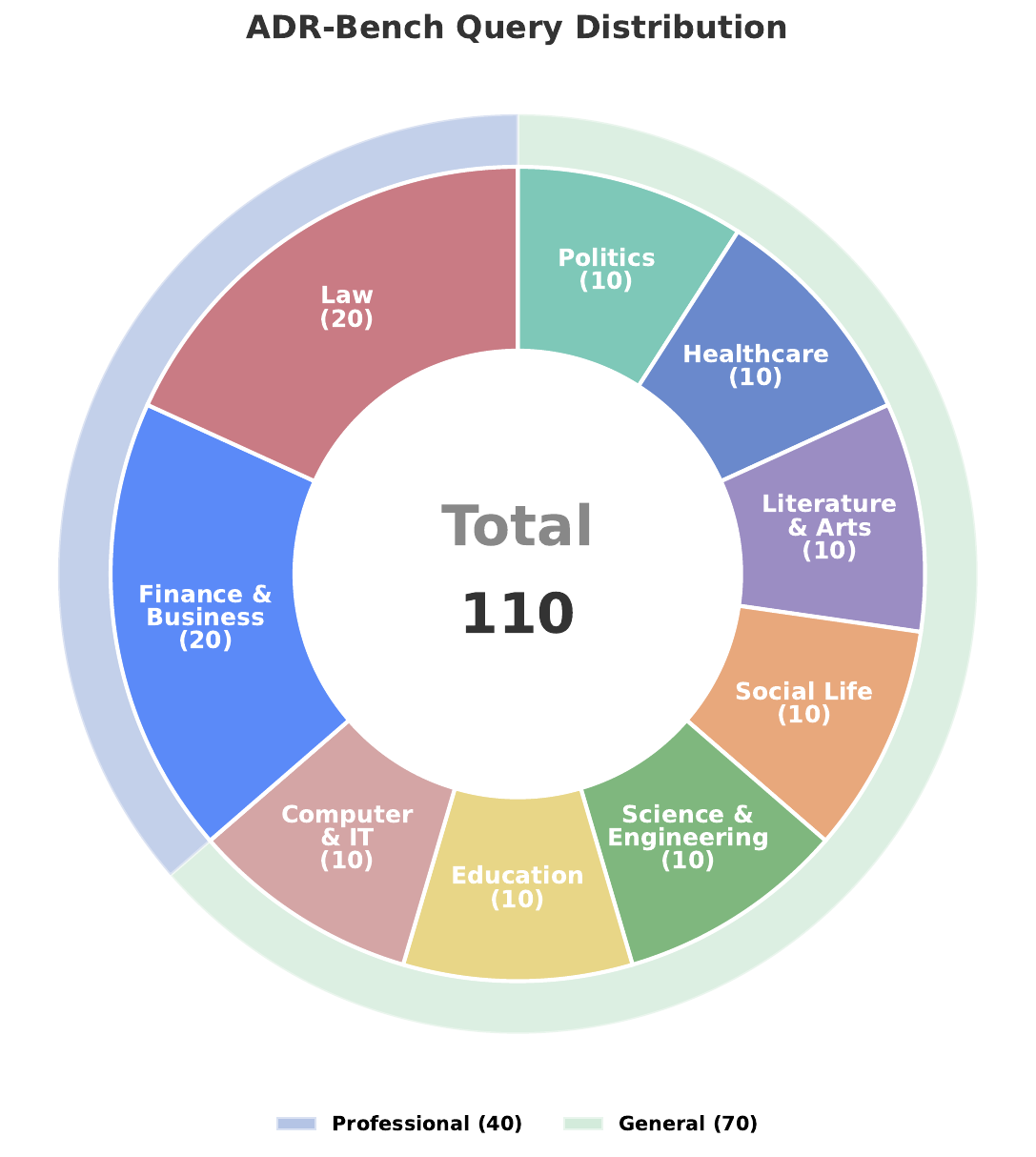}    
    \caption{{The query distribution of ADR-Bench}}
    \label{eval_domain}
\end{figure}

\subsection{Query Composition}
We conducted a comprehensive study of users’ everyday Deep Research usage scenarios and obtained an initial set of queries drafted by domain experts. Through multiple rounds of testing, we filtered and revised queries that were not fully suitable for Deep Research tasks. Based on the distribution of queries in real-world business settings, we categorized user queries into nine domains: Law, Computer and Information Technology, Education, Finance and Business, Science and Engineering, Social Life, Literature and Arts, Healthcare, and Politics.

To construct a more rigorous evaluation framework and better assess the performance of Deep Research, we organized these domains into two primary tracks: general and professional. Law and Finance were selected as representative professional-domain categories. For these, domain experts developed both the queries and their corresponding rubrics, which then underwent rigorous cross-validation. Each of these two domains comprises 20 specialized queries. The remaining seven domains are treated as general-domain tasks, for which we collected authentic queries from real business scenarios. From these, we manually selected high-quality, diverse, and representative samples, assigning 10 queries to each domain. The overall query distribution across these domains is illustrated in Figure \ref{eval_domain}.

\subsection{Evaluation Framework}

We adopt differentiated evaluation strategies for general-domain and professional-domain queries. For general-domain queries, user questions are typically open-ended and broad in scope, with no single standard answer. As a result, it is difficult to objectively characterize report quality through fully enumerated rubrics. Therefore, we employ a human comparative evaluation approach. Specifically, for each task, two reports generated by different models are presented to evaluators under blind-review conditions. Evaluators assign one of five categorical judgments—“Left better / Right better / Both good / Both fair / Both poor”—and score the outputs across four sub-dimensions: information completeness, content depth, requirement fitness, and readability, accompanied by a detailed written justification. Compared with absolute scoring, comparative evaluation introduces a reference anchor, effectively reducing uncertainty stemming from ambiguous quality standards in open-domain scenarios and more accurately capturing the user's subjective perception of report quality. To ensure reliability and consistency, we conducted systematic training for evaluators, standardized both the overall and sub-dimension scoring criteria, and formalized the blind-review procedures. This minimizes the influence of individual preference and enhances the confidence level of evaluation conclusions.

For professional-domain evaluation, we selected Finance and Law as representative domains to assess the model’s capability in generating expert-level reports. Given that these tasks demand specialized domain knowledge and sophisticated reasoning, relying solely on manual comparative review is insufficient for rapid model iteration due to high costs and efficiency constraints. Therefore, we adopt a rubric-based automated evaluation approach: domain experts design representative and relatively constrained real-world tasks, iteratively validate and refine the rubrics through multiple rounds of expert cross-validation. These high-quality, enumerated rubrics ensure comprehensive coverage of the essential elements expected in professional reports, allowing us to rigorously measure the model’s professional competency and knowledge proficiency.

Overall, the proposed evaluation framework captures model-perceived quality differences in real operational contexts for general-domain usage, while leveraging high-quality rubrics in professional domains to monitor professional capability. This approach balances scientific rigor, operational feasibility, and iteration efficiency. Future work will explore more scalable evaluation methodologies to address declining discriminative power and rising evaluation costs as model performance converges.

\subsection{Evaluation Criteria}
Based on the evaluation framework, we conducted a comparative assessment of our model and relevant industry competitors. In this section, we introduce the specific evaluation criteria and the strategic rationale behind their formulation. 

{
\small
\begin{longtable}{>{\columncolor{gray!3.5}}p{0.2\textwidth} p{0.75\textwidth}}
\caption{LLM Response Evaluation Metrics and Guidelines} \label{tab:eval_metrics} \\
\toprule
\rowcolor{headerbg}
\textbf{Dimension} & \textbf{Description \& Criteria} \\
\midrule
\endfirsthead

\multicolumn{2}{c}{{\tablename\ \thetable{} -- continued from previous page}} \\
\toprule
\rowcolor{headerbg}
\textbf{Dimension} & \textbf{Description \& Criteria} \\
\midrule
\endhead

\midrule
\multicolumn{2}{r}{{Continued on next page...}} \\
\bottomrule
\endfoot

\bottomrule
\endlastfoot

\textbf{General \quad\quad\quad\quad Evaluation Logic} & 
\RaggedRight
\textbf{Core Principle:} The score is not a simple summation of sub-dimensions but depends on the query scenario.
\vspace{0.3em}
\begin{itemize}[leftmargin=1.2em, nosep, after=\vspace{0.5em}]
    \item \textbf{Learning/Science Queries:} Prioritize clear logic and gradual deepening (logic > diverse data).
    \item \textbf{Decision/Comparison Queries:} Prioritize hard data comparison and pros/cons analysis (data depth > flowery text).
    \item \textbf{Planning/Proposal Queries:} Prioritize actionable steps and creativity (completeness > complex reasoning).
    \item \textbf{Completeness vs. Depth:} Research queries prioritize completeness, analysis queries prioritize depth.
    \item \textbf{Veto Criteria:} For serious topics (news/history), factual errors or bias result in automatic rejection.
\end{itemize}
\\ \midrule

\textbf{Information \quad\quad Completeness} & 
\RaggedRight
\textbf{Focus:} Breadth, perspective coverage, and avoiding key omissions.
\vspace{0.3em}
\begin{itemize}[leftmargin=1.2em, nosep, after=\vspace{0.5em}]
    \item Does it cover all key aspects? (e.g., market size + competition + policy for industry analysis).
    \item Are obvious or significant pieces of information missing?
    \item For multi-part user prompts, are all sub-questions answered?
\end{itemize}
\\ \midrule

\textbf{Content Depth} & 
\RaggedRight
\textbf{Focus:} Narrative depth, substantial data, evidence, and insight.
\vspace{0.3em}
\begin{itemize}[leftmargin=1.2em, nosep, after=\vspace{0.5em}]
    \item \textbf{Specific Data:} Use of numbers/percentages (e.g., 50 billion market) vs. qualitative descriptions (market is large).
    \item \textbf{Deduction:} Summary and logical deduction vs. simple piling of search results.
    \item \textbf{Insight:} Does it hit critical points and offer valuable conclusions?
\end{itemize}
\\ \midrule

\textbf{Requirement \quad\quad\quad Fitness} & 
\RaggedRight
\textbf{Focus:} Responding to explicit/implicit needs (relevance \& correctness).
\vspace{0.3em}
\begin{itemize}[leftmargin=1.2em, nosep, after=\vspace{0.5em}]
    \item \textbf{Intent:} Does it truly understand the background and deep intent?
    \item \textbf{Constraints:} Are format requests followed? (e.g., output as table or SWOT).
    \item \textbf{Accuracy:} Are key conclusions and data points fact-checked and correct?
\end{itemize}
\\ \midrule

\textbf{Readability} & 
\RaggedRight
\textbf{Focus:} Clarity and friendliness of information organization and presentation. 
\vspace{0.3em}
\begin{itemize}[leftmargin=1.2em, nosep, after=\vspace{0.5em}]
    \item \textbf{Visual Aids:} When presenting complex information (e.g., timelines, comparisons), are appropriate charts used to help users grasp key info quickly?
    \item \textbf{Structure \& Layout:} Are section headings, bolding, and summaries used to make the content hierarchical and scannable? Is the segmentation logical?
    \item \textbf{Accessibility:} Is complex content clarified with key summaries? Are examples used to help readers understand difficult concepts?
\end{itemize}
\\ 
\end{longtable}
}

For comparative evaluations in general domains, a simplified version of our evaluation criteria is shown in Table \ref{tab:eval_metrics}. It is worth noting that the accuracy dimension is difficult to validate reliably in open-ended long-form assessment tasks such as those used for Deep Research, and such validation would impose substantial additional labor and time costs. Therefore, we only require evaluators to verify key conclusions or data points that are directly related to the task requirements and materially affect report quality, and incorporate this requirement under the “alignment with user needs” dimension.

Moreover, because the evaluation dimensions cannot, in practice, be fully disentangled for a given task, evaluators are instructed—when assigning scores on any given dimension—to focus exclusively on the criteria relevant to that dimension without taking other dimensions into account.
For professional domains, we require experts to curate questions with a relatively limited solution space and certain core commonalities. The rubrics must adhere to the five core principles summarized in Table \ref{tab:rubric_principles}.

\noindent

{ 
\small 
\begin{xltabular}{\linewidth}{>{\columncolor{principlebg}\hsize=0.6\hsize\RaggedRight}X >{\hsize=1.4\hsize\RaggedRight}X}
    
    \caption{Principles and Criteria for Constructing Evaluation Rubrics} \label{tab:rubric_principles} \\
    \toprule
    \rowcolor{headerbg}
    \textbf{Principle \& Definition} & \textbf{Examples \& Analysis} \\
    \midrule
    \endfirsthead 

    \multicolumn{2}{c}{{\tablename\ \thetable{} -- Continued from previous page}} \\
    \toprule
    \rowcolor{headerbg}
    \textbf{Principle \& Definition} & \textbf{Examples \& Analysis} \\
    \midrule
    \endhead 

    \midrule
    \multicolumn{2}{r}{{Continued on next page...}} \\
    \endfoot 

    \bottomrule
    \endlastfoot 

    
    \textbf{1. Atomicity} \par \vspace{0.3em}
    Each Rubric must describe only \textit{one clear, single requirement}, assessing whether a specific matter is completed.
    &  
    \begin{itemize}[leftmargin=*, nosep, after=\vspace{0.3em}]
        \item[\cmark] \textbf{Correct:} ``Does the report explain the risk associated with the fund's historical return data?''
        \item[\xmark] \textbf{Incorrect:} ``Does the report detail the statute of limitations, starting point, calculation method, and legal consequences...?''
        \item[] \textcolor{gray}{\textit{\footnotesize $\hookrightarrow$ Analysis: Contains multiple independent requirements.}}
    \end{itemize} 
    \\ \midrule
    
    \textbf{2. Verifiability} \par \vspace{0.3em}
    Rubrics must be actionable with \textit{objective judgment criteria} for explicit verification.
    &  
    \begin{itemize}[leftmargin=*, nosep, after=\vspace{0.3em}]
        \item[\cmark] \textbf{Correct:} ``Does the report cite article XXX of the `civil code' relevant to this case?''
        \item[\xmark] \textbf{Incorrect:} ``Does the report conduct a sufficient and comprehensive discussion...?''
        \item[] \textcolor{gray}{\textit{\footnotesize $\hookrightarrow$ Analysis: ``Sufficient and comprehensive'' is vague and hard to verify.}}
    \end{itemize}
    \\ \midrule
    
    \textbf{3. Unambiguity} \par \vspace{0.3em}
    Phrasing must be clear, ensuring \textit{no multiple interpretations}.
    &  
    \begin{itemize}[leftmargin=*, nosep, after=\vspace{0.3em}]
        \item[\cmark] \textbf{Correct:} ``Does the report list the three major advantages XXX of this policy?''
        \item[\xmark] \textbf{Incorrect:} ``Does the report separately reply to the user's 3 questions?''
        \item[] \textcolor{gray}{\textit{\footnotesize $\hookrightarrow$ Analysis: ``Separately'' is ambiguous (paragraph structure vs. content distinction).}}
    \end{itemize}
    \\ \midrule
    
    \textbf{4. Independence} \par \vspace{0.3em}
    Rubrics should be mutually independent with \textit{no content overlap}.
    &  
    \begin{itemize}[leftmargin=*, nosep, after=\vspace{0.3em}]
        \item[\cmark] \textbf{Correct:} Rubric A checks for securities law citation; Rubric B checks for analysis of key evidence.
        \item[\xmark] \textbf{Incorrect:} Rubric A asks for legal basis citation; Rubric B asks for relevant laws citation.
        \item[] \textcolor{gray}{\textit{\footnotesize $\hookrightarrow$ Analysis: Significant content overlap exists.}}
    \end{itemize}
    \\ \midrule
    
    \textbf{5. Alignment} \par \vspace{0.3em}
    Directly correspond to \textit{core task requirements} without irrelevant dimensions.
    &  
    \begin{itemize}[leftmargin=*, nosep, after=\vspace{0.3em}]
        \item[\cmark] \textbf{Task:} M\&A Legal Report $\to$ \textbf{Rubric:} Identify risks in equity transfer.
        \item[\xmark] \textbf{Task:} Financial Product Risk $\to$ \textbf{Rubric:} Rate company ESG performance.
        \item[] \textcolor{gray}{\textit{\footnotesize $\hookrightarrow$ Analysis: ESG is not a core requirement for this specific task.}}
    \end{itemize}
    \\ 
\end{xltabular}
}

\subsection{Exploratory Attempts}

Throughout the Deep Research evaluation process, we conducted extensive exploratory attempts and gained several insights. For a complex and highly subjective task such as evaluating Deep Research, building a scientific and well-structured evaluation framework is extremely challenging. By sharing our experience, we hope to promote further methodological research on evaluating tasks that are difficult to verify.

\paragraph{Domain segmentation.}
At the early stage of constructing the evaluation framework, we prioritized domain segmentation for Deep Research queries, as this categorization directly informed the composition and distribution of the evaluation set. In our experience, defining too few domains results in insufficient coverage of query distributions, while defining too many domains leads to frequent overlaps, where a query may simultaneously fall into multiple domains. This complicates statistical analysis of query distribution and model performance across domains, reducing statistical meaning and informational value. Ultimately, we found that for Deep Research, defining roughly 6--12 domains was generally reasonable. It is not necessary to pursue a perfectly correct segmentation scheme; however, one guiding principle is that the final segmentation should effectively direct us to the right experts and evaluators for each domain.

\paragraph{Segmentation of evaluation dimensions.}
In comparative evaluations, an overall score helps evaluators express the holistic perceived gap between two reports. Aggregating multi-item results allows us to analyze overall model strengths and weaknesses. However, relying solely on such aggregated scoring only provides the final delivered output of the model, making it difficult to analyze specific issues and guide iteration. Therefore, we consider detailed evaluation dimensions to be necessary. Our experience suggests that each evaluation dimension should be independent, explicit, and executable. Independence means that the dimensions should be as decoupled as possible---though in practice this is difficult, and coupling often appears in some items. Our solution is to evaluate each dimension by focusing only on that dimension at the time. Explicitness means that each dimension should have clear criteria and requirements. Executability means that dimensions must be operational in practice. For example, in the case of long Deep Research reports, verifying each argument or data point is extremely labor-intensive and often unreliable when done manually.

\paragraph{Model-generated rubrics.}
We attempted to use high-performing LLMs such as Gemini to automatically generate rubrics from gold-standard reports, followed by human refinement. Our findings were as follows: (a) the gap between the gold report and an ideal report strongly determines rubric quality. Human revision tends to be influenced by the initial draft, resulting in small adjustments that are hard to validate. Yet obtaining an ideal report is extremely difficult; (b) model-generated rubrics struggle to ensure evaluability when the LLM is used as a judge—for example, verifying data correctness without ground truth, or assessing whether arguments are sufficiently comprehensive; (c) the automatic evaluation results using such rubrics diverge from human evaluators' perception; (d) although the original objective was to let the model initialize rubrics to reduce human workload, in practice substantial rewriting and alignment were still required before the rubrics became usable.

\paragraph{One-sided subjective scoring.}
We attempted a scheme where each query produced three Deep Research reports from different models, and human evaluators assigned 0--10 overall scores. We observed significant variance among evaluators. When restricting the scale to 0--3, the scoring failed to reflect meaningful differences between reports and could not allocate scores reasonably. For Deep Research, where standards are non-unique and unified verification criteria are difficult, comparative evaluation yields more stable results, higher confidence, and improved efficiency.

\paragraph{Resource consumption of Elo battles.}
Elo is a scientifically robust comparative method, but human-based Elo battles require more resources than ordinary comparative evaluations. For example, given 10 queries and 5 models, ordinary comparative evaluation requires only 50 pair comparisons to assess one model against the others. Elo battles require 150 comparisons to produce a full leaderboard. This greatly increases cost for Deep Research, where each comparison is lengthy. Sampling or limiting opponent sets may mitigate the burden, but reduces confidence in the leaderboard while still incurring higher cost than ordinary comparisons. Therefore, for Deep Research, if the goal is simply to measure performance gaps between a model and baseline models, we recommend ordinary comparative evaluation.

\paragraph{Expert-generated questions.}
Senior domain experts who possess an understanding of LLMs and are willing to invest in evaluation work are scarce. Collaborating with such experts incurs significant time and financial costs.


\section{Experiments and Analysis}
\subsection{Experimental Setup}
\paragraph{Evaluation Benchmarks.} We evaluated our model utilizing both established \textsc{ResearchRubrics}~\cite{sharma2025researchrubrics} and our ADR-Bench.
\begin{itemize}
    \item \textbf{LLM-based Evaluation (\textsc{ResearchRubrics})}. We employed LLM judger using a ternary grading for each criterion. Preliminary experiments revealed that the original evaluation prompts exhibited significant stochasticity and failed to accurately interpret negative criteria. To ensure the robustness and reproducibility of our results, we employed targeted prompt engineering for negative constraints in Appendix \ref{app_prompts}, enforced deterministic decoding by setting the temperature to 0, and utilized an ensemble scoring mechanism based on the arithmetic mean of three independent trials per criterion.
    \item \textbf{Human-Centric Evaluation (ADR-Bench)}.
    As detailed in the previous section, we develop two in-house benchmark datasets. The first is a 70-item general test set evaluated via human side-by-side comparison, in which human annotators directly assess reports generated by different models and select the preferred output. This evaluation protocol is designed to capture fine-grained nuances in user preferences and practical utility. The second is a 40-item finance and legal professional test set, where large language models act as evaluators and assign scores based on predefined criteria.
\end{itemize}
\paragraph{Compared models.} We evaluate two representative families of systems.
\begin{itemize}
    \item \textbf{Commercial Agent System.} Since commercial Deep Research agents are closed-source, we manually collected the generated reports from their respective official platforms for evaluation. This category includes OpenAI DeepResearch~\cite{openai2025deepresearch}, Gemini DeepResearch~\cite{google2024gemini}, Kimi-Researcher~\cite{kimi2025researcher}, MiniMax Agent Pro~\cite{minimaxm2}, Qwen DeepResearch~\cite{qwen2025deepresearch}. 
    \item \textbf{ReAct Agent.} 
We implement the ReAct framework across various foundation models via API interfaces. The evaluated models include
Kimi-k2-thinking~\cite{kimiteam2025k2}, DeepSeek-V3.2~\cite{deepseek2025v32}, GLM 4.6~\cite{zeng2025glm45}, MiniMax-M2~\cite{minimaxm2}. Additionally, we introduce our proposed model, Step-DeepResearch, which is initialized from Qwen2.5-32B-Base~\cite{qwen2024tech}. Its development involves a multi-stage training pipeline comprising Mid-training, SFT, and Reinforcement Learning (RL).

\end{itemize} 
\paragraph{Evaluation Configurations.}
To ensure a fair comparison, we standardized the foundation model settings with a maximum of 30 reasoning turns and a limit of 16k tokens per turn. For commercial agent systems, evaluations were performed directly on the final reports generated using their default web-based configurations.

\paragraph{Cost Estimation Standards.}
We adopt a dual-method approach to quantify economic efficiency. For LLM-based ReAct agents, costs are calculated directly from official API pricing and actual token consumption on the \textsc{ResearchRubrics} benchmark. For commercial agent systems without public token statistics, we estimate costs based on the principle that execution time serves as a reasonable proxy for token workload among models of comparable size. Specifically, the cost of MiniMax Agent Pro is estimated by scaling the recorded cost of the MiniMax-M2-based ReAct agent by their execution time ratio. For Kimi-Researcher and Qwen DeepResearch, we utilize the Kimi-k2-thinking-based ReAct agent to establish a baseline, scaling its cost by execution time ratios, given that Qwen3-Max and Kimi-k2 possess similar model parameters. We further adjust the Qwen DeepResearch cost by a factor of $1.6\times$ to reflect the actual API price difference between Qwen3-Max and Kimi-k2.

\subsection{Evaluation Results}

\begin{figure}[ht]
    \centering
\includegraphics[width=\linewidth,
  height=0.3\textheight,
  keepaspectratio]{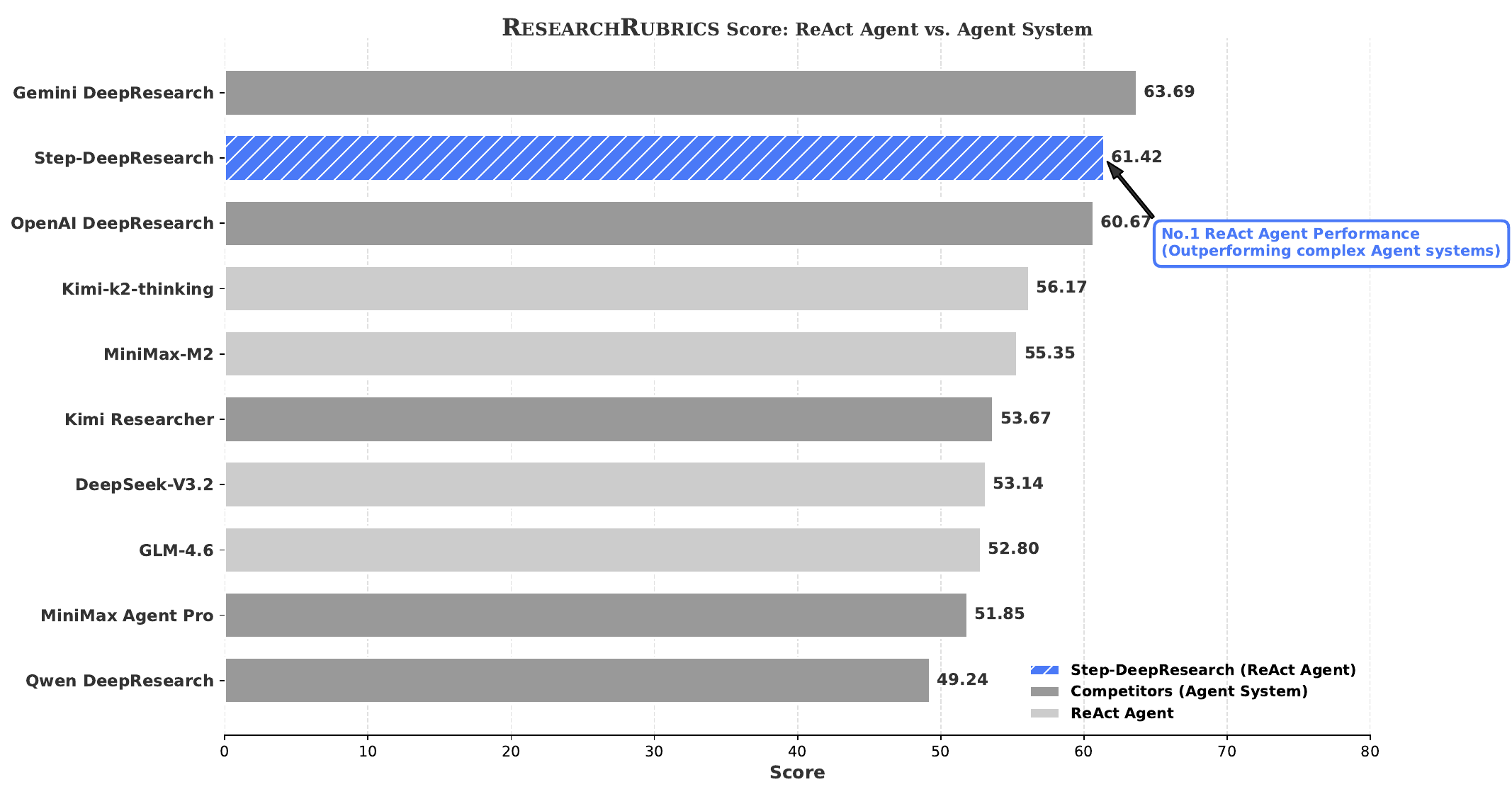}    
    \caption{{Step-DeepResearch Agent against open-source models and commercial products.}}
    \label{model_comparison_rr_total}
\end{figure}

\paragraph{\textsc{ResearchRubrics}.}
Figure \ref{model_comparison_rr_total} presents a comprehensive comparison between Step-DeepResearch and other agents on \textsc{ResearchRubrics} benchmark.
The experimental results demonstrate that Step-DeepResearch achieves state-of-the-art performance within the single-agent category. With a score of $61.42$, it significantly outperforms the leading open-source model, Kimi-k2-thinking ($56.17$), representing a $5.25$ points improvement in performance. In the overall leaderboard, Step-DeepResearch ranks second only to the commercial system Gemini DeepResearch ($63.69$), surpassing most complex agent frameworks including OpenAI DeepResearch and Kimi-Researcher.

Notably, the superior performance of Step-DeepResearch is characterized by exceptional parameter efficiency:
\begin{itemize}
    \item \textbf{Architectural Superiority:} As a single-agent system, its performance exceeds that of various frameworks relying on complex multi-step orchestration or multi-agent collaboration. This validates the model's robust native research capabilities and logical reasoning proficiency.
    \item \textbf{Lightweight Advantage:} Among commercial agent systems, high-end models incur significant costs per report: Gemini DeepResearch ($\approx$ 6.65 RMB) and OpenAI DeepResearch ($\approx$ 5.32 RMB) are the most expensive, followed by MiniMax Agent Pro ($\approx$ 3.36 RMB) and Kimi-Researcher ($\approx$ 2.66 RMB), with Qwen DeepResearch being the outlier ($\approx$ 0.63 RMB).
In the ReAct agent category, costs are generally lower but variable: MiniMax-M2 ($\approx$ 0.42 RMB), DeepSeek-V3.2 ($\approx$ 0.68 RMB), Kimi-k2-thinking ($\approx$ 0.76 RMB), and GLM-4.6 ($\approx$ 1.05 RMB).
Remarkably, Step-DeepResearch achieves a \textsc{ResearchRubrics} score of 61.42 with a single invocation cost of less than 0.50 RMB. At less than one-tenth the expense of top-tier commercial systems such as Gemini and OpenAI, it maintains state-of-the-art performance, demonstrating exceptional cost-effectiveness for large-scale deployment. 
Cost metrics are shown in Figure \ref{fig:cost_perf}.
\end{itemize}

\begin{figure}[ht]
    \centering
\includegraphics[width=0.9\linewidth,
  height=0.35\textheight,
  keepaspectratio]{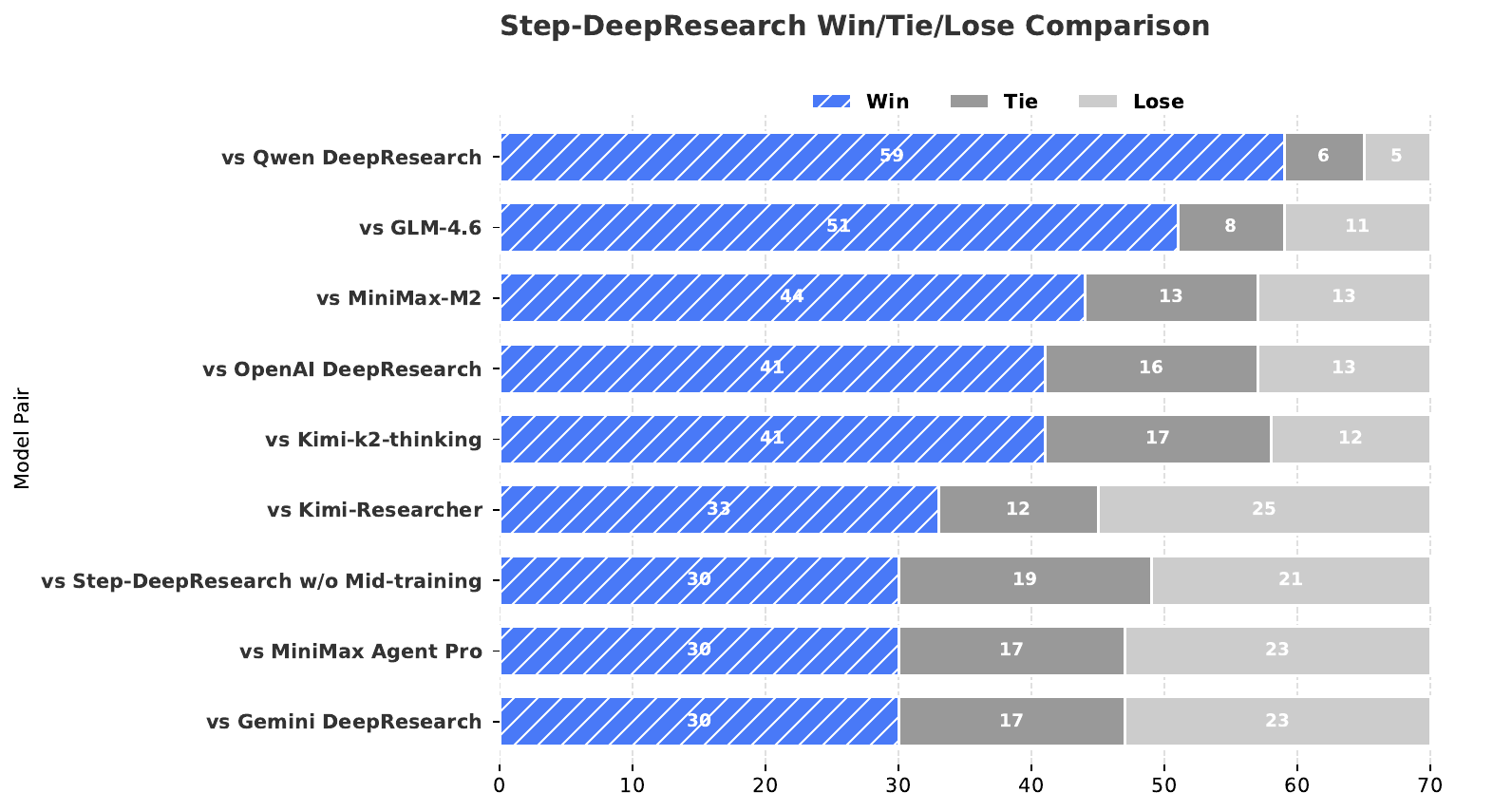}    
    \caption{{\textbf{Human evaluation results on ADR-Bench (N=70)}. The numbers represent the count of Win-Tie-Loss cases for each comparison.}}
    \label{model_comparison_sidebyside}
\end{figure}

\paragraph{ADR-Bench.} 
As shown in Figure \ref{model_comparison_sidebyside}, Step-DeepResearch demonstrates superior performance in human evaluations compared to existing commercial agent systems. In an ablation study comparing Step-DeepResearch with its non-midtrained version, the model achieved a record of $30$ wins and $21$ losses. This result indicates that the integration of mid-training specifically for agents aligns more closely with human preferences regarding the quality of complex research reports. 

When benchmarked against other leading systems, Step-DeepResearch consistently maintains a higher win rate than loss rate, establishing its position as a superior alternative to current systems. Notably, the model demonstrates significant non-inferiority across all  comparisons. Specifically, against formidable opponents such as Gemini and MiniMax, the cumulative count of \enquote{Wins} and \enquote{Ties} reached $47$ ($67.1\%$). These findings provide empirical evidence that Step-DeepResearch consistently meets or exceeds the most advanced performance standards across the vast majority of research scenarios.

\paragraph{ADR-Bench(Finance\&Law)}

 Specifically for the finance and law subsets within ADR-Bench, the tasks are characterized by high-density industry terminology, multi-stage reasoning chains, and strict compliance risk constraints. To address these complexities, we incorporated explicit negative scoring criteria in our checklists to penalize any professional misconceptions identified in the generated reports. For errors deemed fatal or professionally critical by experts, the penalty is particularly severe, leading to a direct \enquote{unusable} (zero-score) rating for that specific report. Under this scoring framework, a model's overall performance is no longer driven solely by its hit rate but also reflects the cost of errors and the externalities of risk. Evaluation results shown in Table \ref{tab:financial_model_performance}, reveal a clear tri-tier distribution of performance. Gemini outperformed all others, establishing a dominant lead, Step-DeepResearch, Kimi, and OpenAI showed comparable performance, clustering within the same tier. The remaining models and products lagged significantly behind, exhibiting a pronounced long-tail effect. These findings suggest that scoring variances are more likely correlated with the models' inherent domain-knowledge coverage rather than gains from agentic frameworks. In other words, under the constraint of strict negative scoring, the process optimizations provided by agent frameworks cannot compensate for a model's fundamental knowledge gaps. Step-DeepResearch's position at the forefront of the second tier is precisely due to its domain-specific training in financial and legal scenarios, granting it a level of expertise competitive with much larger parameter models.

\begin{table}[htbp]
\centering
\caption{Results of ADR-Bench(Finance\&Law) }
\label{tab:financial_model_performance}
\begin{tabular}{ccl}
\toprule
\textbf{Tier} & \textbf{score-range} & \textbf{systems} \\
\midrule
Tier 1 & 25 -- 35 & Gemini DeepResearch \\
\addlinespace
Tier 2 & 15 -- 25 & Step-DeepResearch, Kimi-Researcher, \\
       &              & Kimi-k2-thinking, OpenAI DeepResearch \\
\addlinespace
Tier 3 & 0 -- 15  & Qwen DeepResearch, MiniMax-M2, \\
       &              & MiniMax Agent Pro, GLM-4.6 \\
\bottomrule
\end{tabular}
\end{table}

\subsection{Detailed Analysis}
\paragraph{\textsc{ResearchRubrics}.}

\begin{figure}[h]
    \centering
\includegraphics[width=\linewidth,
  height=0.8\textheight,
  keepaspectratio]{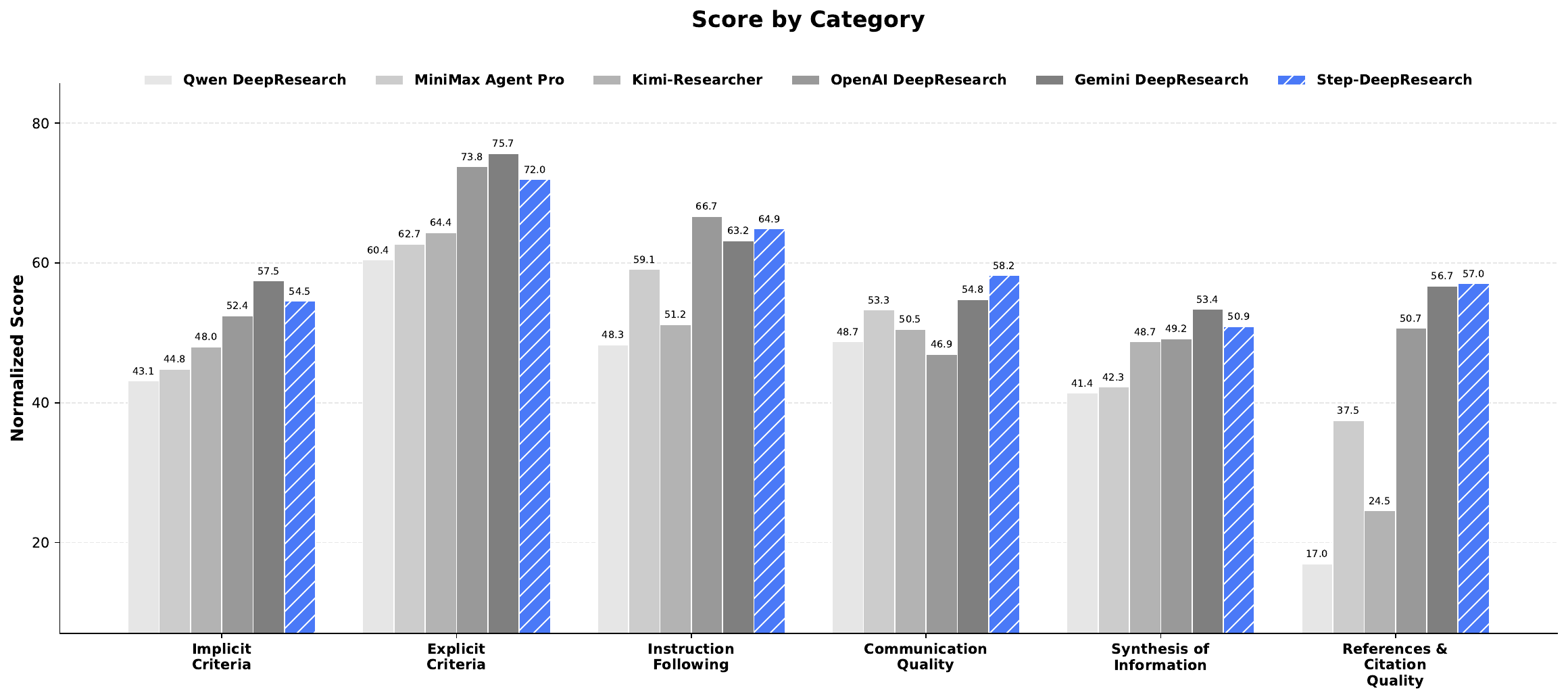}    
    \caption{{ Weighted Score by Category on \textsc{ResearchRubrics}.}}
    \label{rr_category}
\end{figure}

Figure \ref{rr_category}. shows the performance distribution of Step-DeepResearch (denoted by blue diagonal-hatched bars) across six critical evaluation dimensions. The experimental results highlight several key performance breakthroughs:
Step-DeepResearch demonstrated significant leadership in \enquote{Implicit Criteria} and \enquote{Explicit Criteria} securing scores of 54.5 and 72.0, respectively. Notably, its performance in implicit criteria surpasses OpenAI DeepResearch (52.4) and substantially exceeds open-source models, signaling a robust capacity for underlying logical alignment.

In the category of \enquote{Citation Quality} Step-DeepResearch achieved a score of 57.0, tying for the top position with Gemini DeepResearch. This high level of empirical rigor ensures that all generated insights are fully substantiated by verifiable sources. With a score of 58.2 in \enquote{Communication Quality}, the model outperformed all evaluated counterparts. This indicates that the generated reports are not only data-rich but also exhibit superior clarity and professional readability. 
Comparative analysis also identified a marginal gap in \enquote{Instruction Following}, where Step-DeepResearch (64.9) remains slightly behind Kimi-Researcher (66.7). Our preliminary analysis attributes this to the extensive diversity of instruction fine-tuning data during the Post-training phase. Moving forward, we will focus on specialized optimizations for multi-constraint complex tasks to achieve a comprehensive lead across all evaluative dimensions.

\begin{figure}[h]
    \centering
\includegraphics[width=\linewidth,
  height=0.8\textheight,
  keepaspectratio]{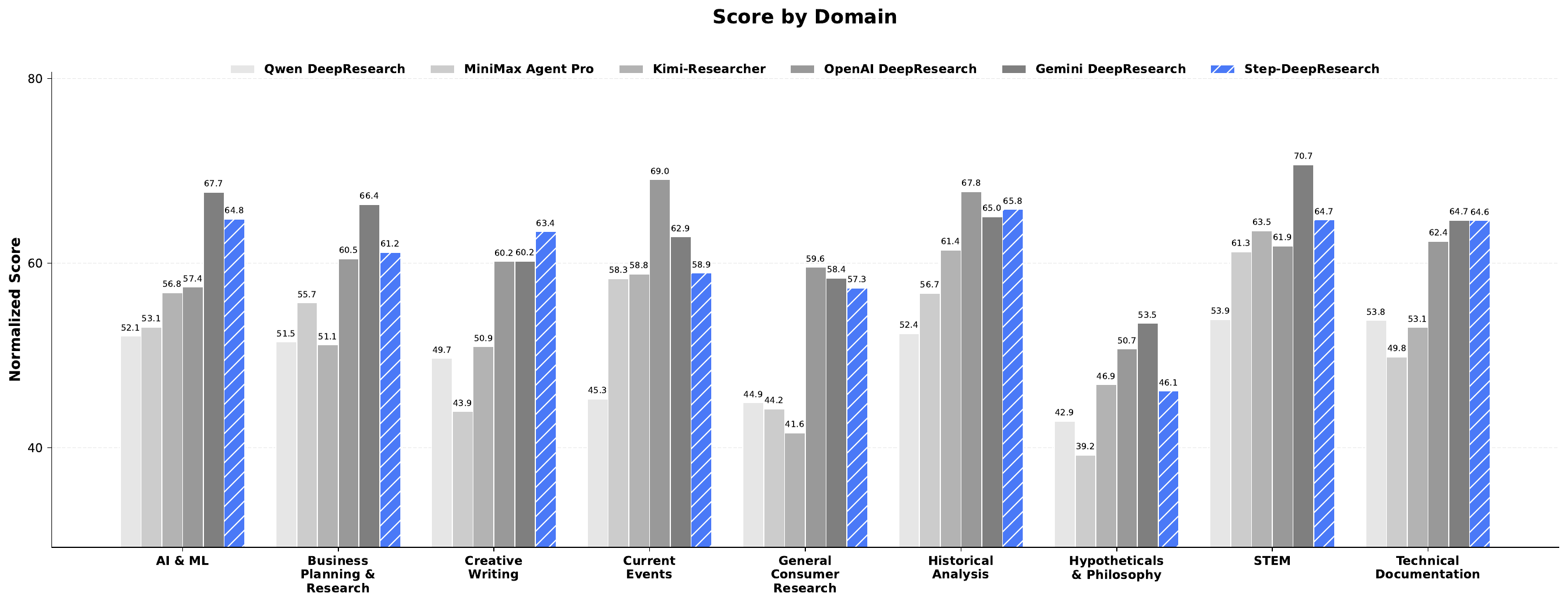}    
    \caption{{Weighted Score by Domain on \textsc{ResearchRubrics}.}}
    \label{rr_domain}
\end{figure}

As shown in Figure \ref{rr_domain}, Step-DeepResearch outperformed Kimi-Researcher in AI \& ML (64.8 vs 57.4), Historical Analysis (65.8 vs 61.4), and Technical Documentation (64.6 vs 53.1), essentially tying with Gemini DeepResearch for the top spot in the latter. Notably, these results were achieved without domain-specific data augmentation. The model also led in Creative Writing (63.4), demonstrating that its underlying logic and knowledge structure generalize effectively to out-of-distribution tasks. Despite broad leadership, the model shows room for improvement in STEM (64.7) and Philosophy (46.1), currently trailing Gemini DeepResearch (70.7 and 53.5, respectively). Error analysis indicates these gaps stem from the inherent complexity of high-order reasoning. Future efforts will prioritize bridging these logical bottlenecks to enhance model performance across STEM and other highly technical domains.

\paragraph{ADR-Bench}

\begin{figure}[ht]
    \centering
\includegraphics[width=\linewidth, keepaspectratio]{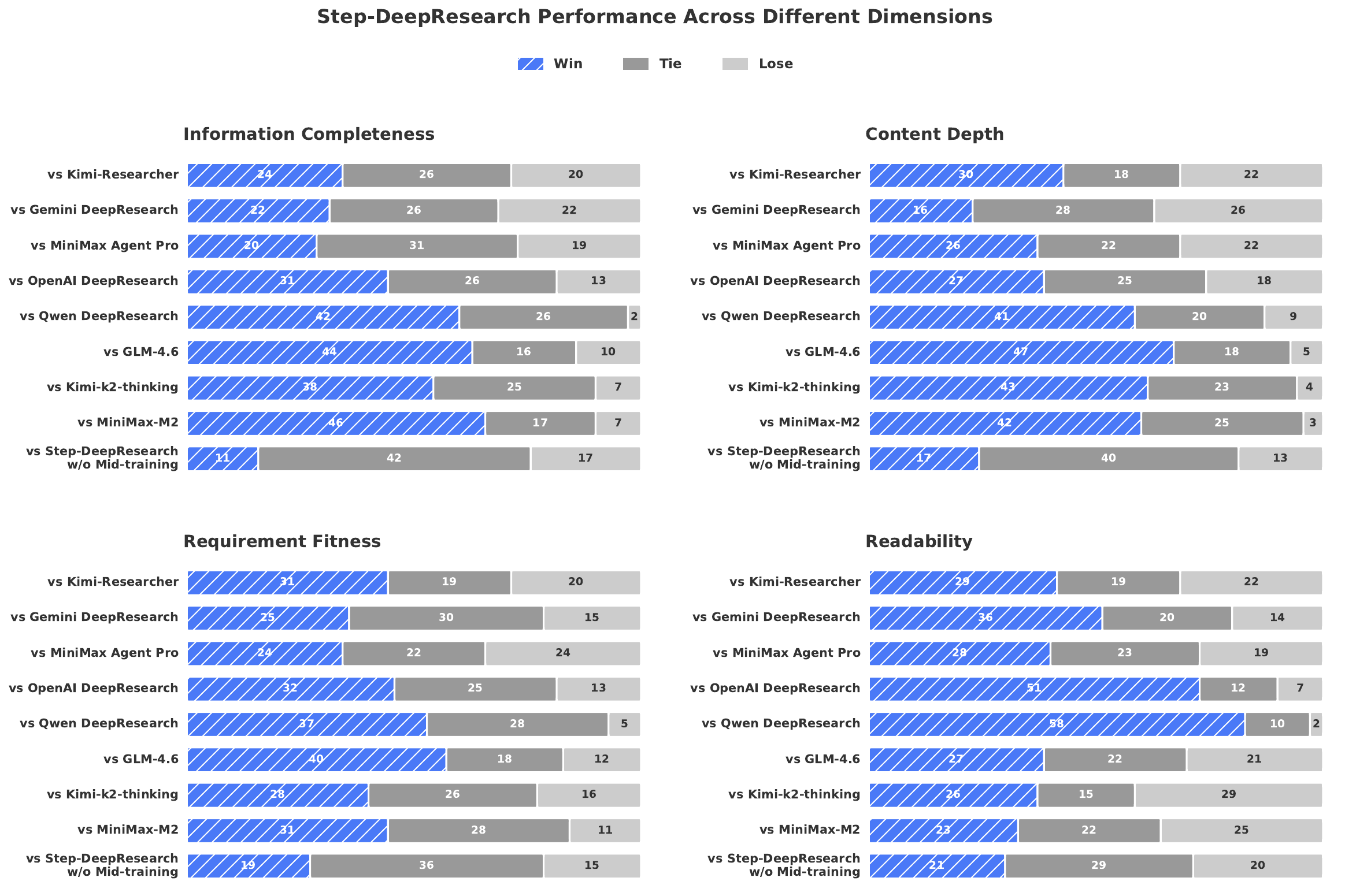}    
    \caption{{Fine-grained results on ADR-Bench}}
    \label{adr-bench-results}
\end{figure}

To assess Step-DeepResearch in general domains, we employed a fine-grained, pairwise human preference evaluation methodology to conduct a comprehensive performance alignment test on Step-DeepResearch, shown in Figure \ref{adr-bench-results}. The following sections present a granular analysis of each performance dimension:
\begin{itemize}
 
\item{\textbf{Informational Completeness.}}
For tasks involving cross-domain knowledge synthesis, evidence chain tracking, and multi-source information aggregation, the model not only covers all key points but also provides a structured presentation, enhancing both information capacity and perceived quality. In tasks requiring large-scale categorization or element listing, the model significantly suppresses content hallucinations and redundant stacking, thereby ensuring a comprehensive and precise output.
\item{\textbf{Content Depth.}}
Regarding content depth, Step-DeepResearch achieves performance comparable to several closed-source commercial systems. In scenarios that heavily rely on specialized professional reasoning, it trails slightly behind commercial Deep Research agents such as OpenAI and Gemini, a performance gap we attribute primarily to constraints in model parameter scale. By contrast, Step-DeepResearch demonstrates more consistent execution capabilities in general problem analysis, planning, and strategic inference tasks. We observe that base models without task-specific training often exhibit shallow response patterns, characterized by short, loosely connected sentences and superficial bullet points that provide limited practical value. Step-DeepResearch effectively rectifies these deficiencies through targeted training.

\item{\textbf{Requirement Fitness.}}
In terms of meeting user requirements, Step-DeepResearch demonstrates a level of stability comparable to that of commercial systems. Across a wide range of instruction complexities, the model is able to accurately identify user intent and provide relevant supplementary information to support decision-making. We observe that several existing commercial systems rely on rigid output patterns, such as imposing a rigid academic structure across diverse tasks or introducing information that is only weakly related to the user’s intent. In contrast, Step-DeepResearch dynamically adapts its presentation style to the specific attributes of the task, ensuring that the output structure remains highly relevant and purpose-driven.
\end{itemize}

\vspace{-1em}
\subsection{Case study}
Table \ref{tab:case-study} presents a representative example from our benchmark comparing the mid-trained model against its non-mid-trained version. The results demonstrate substantial improvements in overall report quality for the mid-trained model. In particular, we observe consistent gains in instruction following, as well as in the breadth and depth of information exploration. These improvements are strongly correlated with the incorporation of atomic capabilities, such as planning and information seeking during the mid-training phase.

\renewcommand{\tabularxcolumn}[1]{m{#1}}
\renewcommand{\floatpagefraction}{0.9} 
\renewcommand{\topfraction}{0.9}       
\renewcommand{\textfraction}{0.1}

\begin{table}[h]
\centering
\caption{Case Study: User Query and Model Comparison}
\label{tab:case-study}
\renewcommand{\arraystretch}{1.5} 
\small
{
\footnotesize
\begin{tabularx}{\textwidth}{|>{\centering\arraybackslash}m{2.5cm}|X|X|}
\hline

\multicolumn{3}{|p{\dimexpr\textwidth-2\tabcolsep-2\arrayrulewidth}|}{\begin{CJK*}{UTF8}{gbsn}
针对大模型的repo级代码工程能力（类似swe bench评测），众多研究提出利用github的pr、issue等数据重构改写数据作为预训练与后训练间的midtrain语料。swe rl给出了数据的详细构造/改写过程，并提出了midtrain（存疑，我不确定它是不是第一个提出代码midtrain概念的）。
我希望了解有没有其他给出github pr、issue数据的详细构造过程，或开源这部分数据清洗重构代码，或开源重构好的数据？我都关注PR/Issue数据的筛选标准、清洗步骤和格式转换方式等特定的数据处理方法。
请给我介绍所有相关工作(如SWE-Fixer、SWE-agent、DeepSWE等)，重点讲解那些对pr、issue以及repo数据组织改写给出详细方案的研究。
\end{CJK*}
} \\

\hline

\multicolumn{3}{|p{\dimexpr\textwidth-2\tabcolsep-2\arrayrulewidth}|}{%
\textbf{English Translation}. Regarding the repo-level code engineering capabilities of large language models (similar to the SWE-bench evaluation), many studies have proposed leveraging GitHub PRs and Issues to construct mid-training corpora between pre-training and post-training phases. SWE-RL provides a detailed data construction/rewriting process and proposes the concept of "mid-training" (though this is questionable, as I am not certain if it was the first to introduce the concept of code mid-training). I am interested in knowing if there are other works that provide detailed construction processes for GitHub PR and issue data, open-source the code for cleaning and reconstructing such data, or provide open-source reconstructed datasets. I am specifically focused on data processing methods, including selection criteria, cleaning steps, and format conversion for PR/Issue data. Please introduce all relevant works (e.g., SWE-Fixer, SWE-agent, DeepSWE), with an emphasis on those that offer detailed schemes for organizing and rewriting PR, Issue, and repo-level data.%
} \\
\hline
\multicolumn{3}{|c|}{\textbf{Model Reports Comparison}} \\
\hline
 \textbf{Dimension} & \textbf{{\footnotesize Step-DeepResearch}} & \textbf{{\footnotesize Step-DeepResearch w/o Mid-training}} \\
\hline
Requirement Fitness &
Fully captures the user's multi-layered requirements, explicitly prioritizing studies that provide detailed PR/Issue and repo-level data construction and rewriting schemes. &
Partially follows the instruction, with limited emphasis on data construction details and weaker alignment with the user's prioritization criteria. \\
\hline
Information Completeness &
Covers a broader range of relevant works, including the SWE-bench series, SWE-smith, SWE-Factory, Multi-SWE-bench, SWE-RL, DeepSWE, InstructCoder, Coeditor, RepoBench, SWE-agent-trajectories, and SWE-Fixer. &
Covers only six works, omitting foundational datasets and key automation frameworks such as SWE-smith and SWE-Factory. \\
\hline
Content Depth &
Provides detailed, end-to-end data processing pipelines, concrete code-level examples, and in-depth discussions of key mechanisms such as Issue-PR pairing, test-case isolation, and automated environment construction. Presents structured comparative tables summarizing data sources, formats, filtering criteria, cleaning steps, format conversions, and open-source availability. &
While providing asymmetrical depth on a select few works (e.g., SWE-RL), the response offers only superficial coverage of the remaining studies. It lacks systematic cross-paper synthesis, presenting information in a fragmented, narrative-driven manner that fails to highlight technical trade-offs.\\
\hline
\end{tabularx}
}
\end{table}

\subsection{Targeted Refinements for Deep Research Patterns}
Based on a systematic analysis of failure cases in earlier iterations, we implemented targeted optimizations for Deep Research scenarios:
\paragraph{Writing Style.} While earlier models achieved high \textsc{ResearchRubrics} scores ($57$), human evaluation revealed a tendency toward fragmented information stacking rather than deep analysis. This highlights a limitation in checklist-based scoring, which prioritizes information stacking rather than deep, integrated analysis. This discrepancy highlights a limitation in checklist-based metrics: they prioritize information recall (i.e., whether key points exist) over structural coherence and narrative depth. Furthermore, we identified a negative correlation between analytical depth and comprehensiveness. When prompted to minimize list-based formatting and provide deeper insights, the model often terminated generation prematurely, leading to a drop in recall. To address this, we optimized our data synthesis pipeline by introducing a Synthesis-driven Drafting module. This module directs the model to transform raw tool-call trajectories into structured paragraphs with logical deduction, while strictly limiting the use of unordered lists as the primary content body. Additionally, we implemented a Pairwise LLM Judger for quality control. This filtering mechanism retains new reports only if they demonstrate superior depth and logic without compromising overall information coverage.
\paragraph{Temporal Cognition.} Despite injecting explicit timestamps into the system prompt, models frequently exhibit temporal confusion during long-horizon reasoning. A notable phenomenon is that models often treat the system-provided date as a \enquote{simulated setting} or habitually append past years (e.g., \enquote{2023} or \enquote{2024}) to search queries—a bias that persists even in synthetic data generated by top-tier closed-source models. Such deviations severely compromise the timeliness and relevance of the final reports. To mitigate this, we implemented strict temporal logic validation during data cleaning: any trajectory where the model anchors a time-agnostic query to a past timestamp is systematically filtered and discarded.
\paragraph{Linguistic Consistency.} To address inconsistent Chinese-English code-switching that disrupts readability, we applied a strict data-cleaning strategy. During the trajectory synthesis phase, we utilize regular expressions and language density detection to exclude low-quality trajectories characterized by unnecessary language mixing, ensuring a consistent and fluent output.


\section{Conclusion and Future Work}
In this work, we have revisited the meta-capabilities of Deep Research agents and proposed a holistic understanding of the Deep Research task, achieving superior performance on a 32B-parameter model. Specifically, by leveraging a multi-stage atomic-capability data strategy and an end-to-end training paradigm, we have progressively improved the model's research proficiency from agentic mid-training to post-training. Furthermore, we have introduced \textbf{ADR-Bench}, a novel benchmark designed to evaluate the practical usability of agents in real-world scenarios. Extensive experiments demonstrate that our principles and paradigms achieve state-of-the-art results among medium-sized models and can compete with proprietary large-scale models in specific domains. We hope this work will inspire the community to further advance the frontier of autonomous agents toward AGI.

Despite these advancements, we currently face several challenges. First, the generalization and robustness of tool use remain insufficient; the system often exhibits brittle points when encountering API variations, anomalous returns, or long-chain tasks involving complex cross-tool compositions. Second, ensuring stable overall correctness and factuality remains difficult, particularly in scenarios with high information noise or fragmented evidence, which can lead to ``plausible but unprovable'' inferences. Finally, the readability and auditability of the generated reports have room for improvement, such as more consistent structural organization and more explicit mapping between conclusions and evidence.

Our future research aims to address these challenges through three strategic advancements. Initially, we focus on collaborative intelligence, introducing a multi-agent paradigm where specialized roles—planners, retrievers, verifiers, and writers—reduce hallucinations via consensus mechanisms. Beyond architectural design, we seek to enhance environmental adaptability, enabling agents to perform continuous exploration and error correction in dynamic, partially observable settings. Finally, we prioritize the rigor of training objectives by incorporating metrics such as factual consistency and traceable citations. Through the synergy of preference learning and process-based supervision, we strive to ensure that model outputs are not only ``seemingly correct'' but ``verifiably and clearly correct.''

\section{Contributors and Acknowledgements}

This work was carried out by the \textbf{Agent Team} at \textbf{StepFun}. 
Within each contribution category, contributors are listed in alphabetical order by \emph{first name}.

\paragraph{Corresponding Contributors.}
Chen Hu (\texttt{hatcher@stepfun.com}), Daxin Jiang (\texttt{djiang@stepfun.com}).

\paragraph{Research and Data.}
Chen Hu, Haikuo Du, Heng Wang, Lin Lin, Mingrui Chen, Peng Liu, Ruihang Miao, Tianchi Yue, Wang You, Wei Ji, Wei Yuan, Wenjin Deng, Xiaojian Yuan, Xiaoyun Zhang, Xiangyu Liu, Xikai Liu, Yanming Xu, Yicheng Cao, Yifei Zhang, Yongyao Wang, Yubo Shu, Yurong Zhang, Yuxiang Zhang, Zheng Gong and Zhichao Chang.

\paragraph{Evaluation.}
Binyan Li, Dan Ma, Furong Jia, Hongyuan Wang, Jiayu Liu, Jing Bai, Junlan Liu, Manjiao Liu, Na Wang, Qiuping Wu, Qinxin Du, Shiwei Li, Wen Sun, Yifeng Gong, Yonglin Chen, Yuling Zhao, Yuxuan Lin, Ziqi Ren and Zixuan Wang.

\paragraph{Infrastructure.}
Aihu Zhang, Brian Li, Buyun Ma, Kang An, Li Xie, Mingliang Li, Pan Li, Shidong Yang, Xi Chen, Xiaojia Liu, Yuchu Luo, Yuan Song, YuanHao Ding, Yuanwei Liang, Zexi Li, Zhaoning Zhang and Zixin Zhang.

\paragraph{Project Sponsors.}
Binxing Jiao, Daxin Jiang, Jiansheng Chen, Jing Li, Xiangyu Zhang and Yibo Zhu.

\section*{Acknowledgements}

We would like to express our sincere gratitude to the following colleagues at \textbf{StepFun} for their support:

Alex Chen, Bingxin Li, Beidi Luan, Chang Su, Dionysia Zhang, HuanCheng Bai, Huangxi Zhu, Jiacan Dong, Junhan Gu, Junjing Guo, MingMin Li, Qianyu Yang, Qi Zhou, Ran Ding, Ran Sun, Rui Sun, Shanshan Yuan, Sitong Liu, Tianyi Zhang, Weibo Wu, Wenjing Zhao, Xuan He, Yichen Wang, Yixuan Kong, Zephyr Zhu, Zixuan Chen and Zhixin Chen.

\bibliographystyle{unsrtnat}
\bibliography{refs}

\clearpage

\appendix
\section*{Appendix}
\renewcommand{\thesubsection}{\Alph{subsection}.}
\subsection{\textsc{ResearchRubrics} Judger System Prompt}\label{app_prompts}

\begin{tcblisting}{
    enhanced,
    breakable,
    title=\textsc{ResearchRubrics} Judger System Prompt, 
    colback=gray!5, 
    colframe=black!75, 
    fonttitle=\bfseries,
    listing only,
    listing options={
        breaklines=true,
        basicstyle=\ttfamily\fontsize{8.5pt}{8.5pt}\selectfont,
        breakatwhitespace=false,
        keywords={PositiveRubricsSystemPrompt, NegativeRubricsSystemPrompt, USERPrompt},
        keywordstyle=\color{red}\bfseries,
        % ---------------------------------------
    }
}
PositiveRubricsSystemPrompt
You are an expert evaluator tasked with assessing whether a document satisfies specific rubric criteria. Your evaluation must be precise, objective, and based solely on the evidence present in the document.
## Evaluation Framework
You will evaluate each rubric criterion using a three-tier satisfaction scale:
1. **Not Satisfied (Score: 0.0)**: The document fails to meet the criterion. Key elements are missing, incorrect, or inadequately addressed.
2. **Partially Satisfied (Score: 0.5)**: The document partially meets the criterion. Some elements are present but incomplete, lacking depth, or missing important aspects.
3. **Satisfied (Score: 1.0)**: The document fully meets the criterion. All required elements are present, well-developed, and appropriately detailed.
## Evaluation Process
1. **Understand the Criterion**: Carefully read and interpret what the rubric is asking for.
2. **Search for Evidence**: Systematically review the document for relevant content that addresses the criterion.
3. **Assess Completeness**: Evaluate whether the evidence fully, partially, or fails to satisfy the criterion.
4. **Provide Reasoning**: Explain your evaluation with specific references to the document content.
## Important Guidelines
- Base your evaluation ONLY on what is explicitly present in the document
- Do not make assumptions about implied or missing content
- Consider the quality, completeness, and relevance of the evidence
- Be consistent in your evaluation standards across all criteria
- Provide specific examples from the document to support your verdict
Note: Example lists in these rubrics are intended to illustrate possible reasoning patterns or relevant topics. These example lists contain correct answers but are not exhaustive. Use them as guidance, but also make your own final judgment about what qualifies as correct when appropriate.

NegativeRubricsSystemPrompt
You are an expert Quality Control (QC) evaluator tasked with assessing whether a document contains a specific **negative attribute or flaw** defined by the rubric criterion. Your evaluation must be precise, objective, and solely focused on the presence and extent of this flaw.
**Note:** The criterion you are evaluating is a defect; your score should reflect the degree to which this defect is present in the document. A score of 0.0 indicates a high-quality document (it successfully *avoids* the flaw), and a score of 1.0 indicates a low-quality document (it fully *exhibits* the flaw).

## Evaluation Framework: Flaw Detection Scale
You will evaluate the criterion based on the *presence* of the negative attribute, using the following scale:
1. **Not Satisfied (Score: 0.0)**: **The Flaw is Absent.** The document successfully avoids the described negative attribute. On this aspect, the document is high-quality.
2. **Partially Satisfied (Score: 0.5)**: **The Flaw is Partially Present.** The document partially exhibits the negative attribute, or the flaw is minor, isolated, or inconsistent.
3. **Satisfied (Score: 1.0)**: **The Flaw is Fully Present.** The document clearly, consistently, and substantially exhibits the described negative attribute. On this aspect, the document is low-quality.

## Evaluation Process
1. **Identify the Flaw**: Carefully read the criterion to understand the exact negative attribute (flaw) being defined (e.g., "harsh tone," "bias," "missing key data"). Note that this criterion typically has a negative 'Weight'.
2. **Search for Evidence of the Flaw**: Systematically review the document for specific content that demonstrates the presence of this negative attribute.
3. **Assess Extent of Flaw**: Evaluate whether the flaw is absent (0.0), partial (0.5), or fully present (1.0).
4. **Provide Reasoning**: Explain your evaluation with specific references to the document content, focusing on *where* and *how often* the flaw appears.

## Important Guidelines
- **Flaw Focus**: Your entire focus is on detecting the presence of the defined flaw.
- **Inverted Quality Logic**: Remember that a low score (0.0) means the document is *good* because it avoided the flaw, and a high score (1.0) means the document is *bad* because the flaw is present.
- Base your evaluation ONLY on what is explicitly present in the document.
- Provide specific examples from the document to support your verdict, especially quotes that demonstrate the flaw (if present).

\end{tcblisting}

\subsection{Examples}\label{app:examples}

\begin{longtable}{@{} >{\columncolor{gray!3.5}}p{0.12\textwidth} p{0.84\textwidth} @{}}
\caption{Prompt Examples across Five Distinct Domains} \label{tab:domain_examples} \\
\toprule
\rowcolor{headerbg}
\textbf{Category} & \textbf{Chinese Prompt} \\
\midrule
\endfirsthead

\multicolumn{2}{c}%
{{\tablename\ \thetable{} -- continued from previous page}} \\
\toprule
\rowcolor{headerbg}
\textbf{Category} & \textbf{Chinese Prompt} \\
\midrule
\endhead

\midrule
\multicolumn{2}{r}{{Continued on next page}} \\
\bottomrule
\endfoot

\bottomrule
\endlastfoot

\textbf{Social Life} & 
\begin{CJK*}{UTF8}{gbsn}
\textbf{Scenario:} 多约束复杂旅行规划：带娃、多人、出国、城市移动旅行方案。\newline
\textbf{Prompt:} 我们一行5人（爷爷奶奶爸爸妈妈和一个两岁半的宝宝），准备在明年十一假期(9月30日-10月8日)去日本玩，从上海往返。帮我设计全程8天的旅行方案。
\begin{itemize}
    \item 我想要福冈进，鹿儿岛出，中间也去熊本，不要走回头路（暂时的计划，如果不合理你告诉我）
    \item 整个行程不要太累，以体验自然风光和享受美食为主
    \item 住宿正常安排，考虑交通便利性和酒店舒适度
    \item 公共交通为主
    \item 餐饮选择要考虑到有小朋友，不太喜欢寿司刺身之类的冷盘，最好有热食
    \item 温泉可以考虑，但非必须，综合所有行程看，合适的话可以安排
    \item 预算范围正常区间即可，不用过度节约或太奢侈 
\end{itemize}
\end{CJK*} \\
\midrule

\textbf{Science \& Engineering} & 
\begin{CJK*}{UTF8}{gbsn}
\textbf{Scenario:} 企业级AI基础设施规划：大规模推理、容量预测与精细化成本治理。\newline
\textbf{Prompt:} 我有一个大规模的对客户提供API接口服务的开放平台，和相应的LLM及多模态AI推理服务。为了更好地提升资源利用率和做好对客户的服务保障，我希望建立模型部署资源测算与管理机制，特别关注性能保障、容量预估以及成本分析和控制方面。我的平台有明显的高峰期，用户主要分布在国内。作为专业的架构师，请写出你的最终设计方案。
\end{CJK*} \\
\midrule

\textbf{Politics} & 
\begin{CJK*}{UTF8}{gbsn}
\textbf{Scenario:} 全域地缘政治画像：欧盟27国立场分类、历史演进与产业动因分析。\newline
\textbf{Prompt:} 帮我梳理欧盟各国在欧盟人工智能治理问题上的立场，客观阐述，以事实为主，不要遗漏，也不要过度上升。从他们最早讨论AI到现在，需要涵盖所有成员国，按照他们的态度做分类，并且深挖不同态度的成因，比如是否和自身技术、产业发展有关）
\end{CJK*} \\
\midrule

\textbf{Finance \& Business} & 
\begin{CJK*}{UTF8}{gbsn}
\textbf{Scenario:} 行业周期复盘：多年数据回溯、增长/衰退根因分析、风口与风险研判。\newline
\textbf{Prompt:} 你是一名资深数字经济领域研究员，请结合国内平台治理节奏和竞争格局变化的背景，复盘直播电商（2019-2025E）市场规模变迁并归因。完成以下任务：1）指出行业“遇冷”的拐点，依据数据表现（比如同比转负）界定下行区间；基于这些原因，总结未来导致直播电商规模继续萎缩或再次下滑的主要风险是什么，并说明这种风险一般会出现在什么样的市场环境里。2）找出高增速的年份，说明那一年为什么变热；基于这些原因，总结未来可能推动直播电商规模重回增长轨道的机会，并说明这些机会成立需要满足哪些前提条件。
\end{CJK*} \\
\midrule

\textbf{Law} & 
\begin{CJK*}{UTF8}{gbsn}
\textbf{Scenario:} 电商平台合规诊断：支付二清研判、发票流转合规性。\newline
\textbf{Prompt:}
\begin{enumerate}
    \item A商城是一家为注册用户提供线上交易撮合服务的电商平台，A商城的业务模式为：买家通过A商城购买卖家的产品后，货款先转入A商城在银行开立的资金托管账户，在买家验收或其他付款条件达成后，该资金托管账户进行清分，其中货款的千分之五作为交易佣金（技术服务费）支付到A商城自有账户，剩余货款支付到卖家自有账户。
    \item A商城与B银行签署了相关协议，B银行为A商城开立资金存管账户，该账户不属于A商城所有，B银行按照A商城的指令向A商城及卖家办理资金结算（向A商城支付交易佣金，向卖家支付货款），
    \item 卖家向买家开具货款发票、A商城向卖家开具佣金发票。
\end{enumerate}
A商城的上述业务模式是否存在税务及法律方面的合规风险？
\end{CJK*} \\

\end{longtable}

\end{document}